\documentclass[lettersize,journal]{IEEEtran}
\usepackage{amsmath,amsfonts}
\usepackage{algorithmic}
\usepackage{algorithm}
\usepackage{array}
\usepackage[caption=false,font=normalsize,labelfont=sf,textfont=sf]{subfig}
\usepackage{textcomp}
\usepackage{stfloats}
\usepackage{url}
\usepackage{verbatim}
\usepackage{graphicx}
\usepackage{cite}
\usepackage{amsfonts,amssymb}
\newcommand{\bs}{\boldsymbol}
\usepackage{color}
\usepackage{multirow}
\usepackage{booktabs}
\usepackage{bbding}
 \usepackage[colorlinks,linkcolor=red]{hyperref}
\begin{document}

\title{3D-OAE: Occlusion Auto-Encoders for Self-Supervised Learning on Point Clouds}

\author{Junsheng Zhou\IEEEauthorrefmark{1}, 
        Xin Wen\IEEEauthorrefmark{1},
        Baorui Ma,
        Yu-Shen Liu,
        Yue Gao, 
        Yi Fang, 
        Zhizhong Han
        
\thanks{Junsheng Zhou and Xin Wen contribute equally to this work.}
\thanks{Junsheng Zhou and Baorui Ma are with the School of Software, Tsinghua University, Beijing, China (e-mail: zhoujs21, mbr18@mails.tsinghua.edu.cn).}
\thanks{Xin Wen is with JD Logistics, JD.com, Beijing, China (email: wenxin16@jd.com).}
\thanks{Yu-Shen Liu and Yue Gao are with the School of Software, BNRist, Tsinghua University, Beijing, China (e-mail: liuyushen@tsinghua.edu.cn, kevin.gaoy@gmail.com). Yu-Shen Liu is the corresponding author.}
\thanks{Yi Fang is with New York University, New York, USA (email: yfang@nyu.edu).}
\thanks{Zhizhong Han is with the Department of Computer Science, Wayne State University, USA (e-mail: h312h@waume.edu).}
\thanks{This work was supported by the National Natural Science Foundation of China (62272263, 62072268), and in part by Tsinghua-Kuaishou Institute of Future Media Data. Project page at \url{https://junshengzhou.github.io/3D-OAE}.}
}



\maketitle

\begin{abstract}
The manual annotation for large-scale point clouds is still tedious and unavailable for many harsh real-world tasks.
Self-supervised learning, which is used on raw and unlabeled data to pre-train deep neural networks, is a promising approach to address this issue. Existing works usually take the common aid from auto-encoders to establish the self-supervision by the self-reconstruction schema. However, the previous auto-encoders merely focus on the global shapes and do not distinguish the local and global geometric features apart. To address this problem, we present a novel and efficient self-supervised point cloud representation learning framework, named 3D Occlusion Auto-Encoder (3D-OAE), to facilitate the detailed supervision inherited in local regions and global shapes. We propose to randomly occlude some local patches of point clouds and establish the supervision via inpainting the occluded patches using the remaining ones. Specifically, we design an asymmetrical encoder-decoder architecture based on standard Transformer, where the encoder operates only on the visible subset of patches to learn local patterns, and a lightweight decoder is designed to leverage these visible patterns to infer the missing geometries via self-attention. We find that occluding a very high proportion of the input point cloud (e.g. 75$\%$) will still yield a nontrivial self-supervisory performance, which enables us to achieve 3-4 times faster during training but also improve accuracy. Experimental results show that our approach outperforms the state-of-the-art on a diverse range of downstream discriminative and generative tasks.
\end{abstract}


\section{Introduction}
\IEEEPARstart{P}{oint} clouds play a crucial role in 3D computer vision applications \cite{cui2021deep,alexiou2017towards,pomerleau2015review} due to its flexibility to represent arbitrary geometries and memory-efficiency. In this paper, we specifically focus on the task of learning representations of point clouds without manually annotated supervision. 
As 2D images, learning representations for 3D point clouds has been comprehensively studied for many years, and the research line along the 2D and 3D representation learning shares a lot of common practices, such as the auto-encoder based framework and the self-reconstruction based supervision. The recent development in both NLP and 2D computer vision fields has also driven several improvements in 3D representation learning, such as PCT \cite{guo2021pct}, Point-BERT \cite{yu2021pointbert} and STRL \cite{huang2021spatio}. However, the different data characteristics between the 2D and 3D domains limit the direct applications of many 2D improvements into 3D scenarios, e.g. the differences between ordered 2D grids and unordered 3D points.

One of major challenges is to learn the hierarchical context between the global structure and local geometries. In 3D scenarios, this is more difficult than the learning of 2D images due to the discrete nature of 3D points. In most previous methods, the 3D auto-encoder usually relies on the self-reconstruction as the supervision to focus on the global structures and the local parts. However, the simple self-reconstruction based framework usually does not explicitly distinguish the local parts and the global structures apart. As a result, both of them are only revealed by the shape matching constraints (e.g. Chamfer Distance) as a whole, while more detailed self-supervision to reveal the local to global hierarchy in 3D point clouds is merely discussed.

\begin{figure*}[!t]
  \centering
  \includegraphics[width=2\columnwidth]{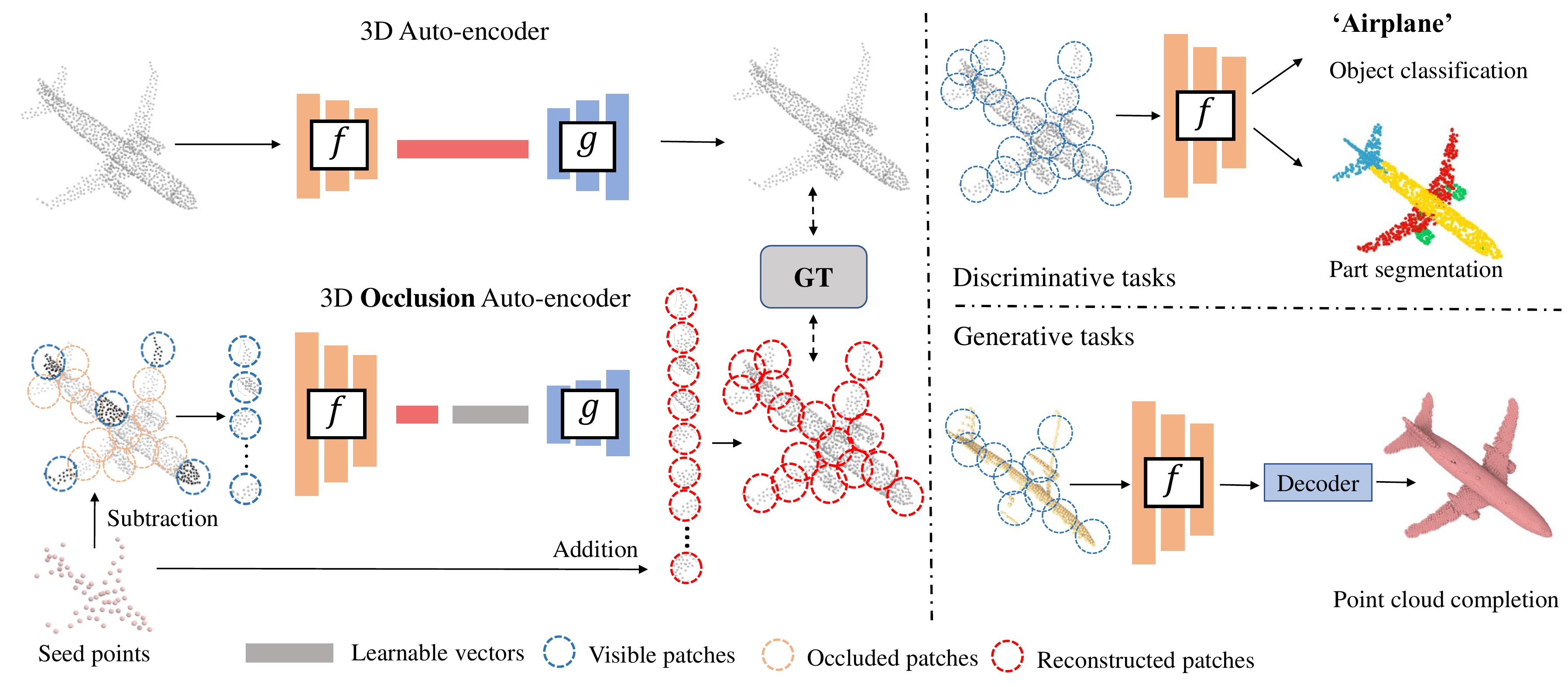}
  \caption{\textbf{Comparison between previous auto-encoders and our 3D-OAE.} The $f$ and $g$ indicate the encoder and decoder of an auto-encoder. The seed points is extracted from the original shape using FPS. Unlike other auto-encoders which takes the whole shape as input to reconstruct itself, 3D-OAE randomly occludes a high ratio of patches, encodes only on the visible patches, and learns to recover the complete shape. After self-supervised learning, we keep $f$ for further fine-tuning. We demonstrate that the $f$
learned by 3D-OAE shows powerful performances in both discriminate tasks (e.g. object classification, part segmentation) and generative tasks (e.g. point cloud completion).}
  \label{fig:3D-OAE}
\end{figure*}

The recent improvements of mask-based 2D auto-encoders \cite{he2021masked} have proved that masked auto-encoders are effective in image representation learning through the inference of the overall image information based on the visible local patches. It provides a new perspective to establish the self-supervision between the local and global information. 
However, due to the discrete nature of point clouds, it's difficult to directly use a 2D mask-based auto-encoder to learn 3D representations.
Driven by the above analysis, we present 3D-OAE, a novel Transformer-based self-supervised learning framework with Occlusion Auto-Encoder. As shown in Fig. \ref{fig:3D-OAE}, we separate an unlabelled point cloud into local point patches and centralize them to their corresponding seed point. After that, we occlude a large proportion of the patches but remain the seed points, and learn to recover occluded patches from seed points and the visible patches. The seed points serve as global hints to guide the shape generation and the model will be forced to focus on learning the local geometries details.
Specifically, we design an encoder to learn features only on the visible subset of patches, and a decoder to leverage the features of visible patches to predict the local features of the occluded ones, and finally reconstruct the occluded patches with seed points as the global hints.
After self-supervised learning without any manual annotation, we can transfer the trained encoder to different downstream tasks. We demonstrate our superior performances by comparing our method under widely used benchmarks.

Our main contributions can be summarized as follows:
\begin{itemize}
    \item We proposed a novel self-supervised learning framework named 3D Occlusion Auto-Encoder. Unlike previous 3D auto-encoders, 3D-OAE designs an asymmetrical encoder-decoder Transformer architecture to learn the patterns from the visible local patches and leverage them to control the local geometry generation of the occluded 
    patches. After self-supervised learning, the trained encoder can be transferred to new downstream tasks.
    \item Our 3D-OAE can remove a large proportion (e.g. 75$\%$) of point cloud patches before training and only encodes a small number of visible patches. This enable us to accelerate training for 3-4 times and makes it possible to do self-supervised learning in large scale unlabelled data efficiently.
    \item We achieved the state-of-the-art performances in six different downstream applications compared with previous self-supervised methods.
    
\end{itemize}

\section{Related Work}
\subsubsection{Representation Learning on Point Clouds}
Different from structured data like images, point clouds are unordered sets of vectors, which brings great challenges to the learning of representations. The deep learning based 3D point cloud processing techniques has achieved very promising results in different tasks \cite{ma2021neural,chen2021unsupervised,li2022learning,3DAttriFlow,On-SurfacePriors,PredictableContextPrior,spunet,pmp++, Zhou2022CAP-UDF,sheng2021deep,akhtar2021video,qiu2021geometric}.  
Qi et al. \cite{qi2017pointnet} pioneered point cloud learning by proposing PointNet, which directly input the raw point cloud into point-wised MLPs and use max-pooling to solve the permutation invariance of point cloud. Further more, PointNet++ \cite{qi2017pointnet++} applies query ball grouping and hierarchical spatial structure to increase the sensitivity to local geometries. Some works followed this idea and developed different grouping strategies \cite{wu2019pointconv,yang2019modeling,hu2020randla,zhang2020pointhop}. A number of approaches build graphs to connect points and aggregate information through graph edges \cite{wang2019dynamic,gadelha2018multiresolution,verma2018feastnet,shen2018mining,wang2019graph,chen2020hapgn}. DGCNN \cite{wang2019dynamic} was proposed to use graph convolutions on KNN graph nodes, and GACNet \cite{wang2019graph} apply attention mechanism in graph convolutions. Some other methods were proposed to use continuous convolutions on point clouds\cite{hua2018pointwise,xu2018spidercnn,zhang2019shellnet,su2018splatnet,li2018pointcnn,wu2019pointconv,thomas2019kpconv}. PointCNN \cite{li2018pointcnn} applies convolution neural networks on point set after reordering points with special operators. 
Different from the idea of using convolution-based structure, Point2Sequence \cite{liu2019point2sequence} was proposed to use a recurrent model (i.e. RNN) to capture the fine-grained sequence information of features in local patches. Point2SpatialCapsule \cite{wen2020point2spatialcapsule} introduces a capsule network for modeling the spatial relationships of local regions by aggregating the local features into a set of learnable cluster centers.

Some recent studies focus on improving the extraction of local features. RS-CNN \cite{liu2019relation} proposes a shape-aware convolution network to learn local features by exploring the relationship of local points. ShellNet \cite{zhang2019shellnet} designs a permutation invariant convolution operation for learning and aggregating the geometric features of local regions with different scales. GS-Net \cite{xu2020geometry} was proposed to use the Eigen-Graph to learn the geometric relationships between local point cloud regions. A more recent work Point-MLP \cite{ma2021rethinking} introduces a MLP-based network with a learnable geometric affine operation to adaptively transform the point features in local regions.

Inspired by the great success of Transformers in both NLP \cite{vaswani2017attention,devlin2018bert,joshi2020spanbert} and 2D vision \cite{carion2020end,liu2021swin,dosovitskiy2020image,bao2021beit}, some recently works try to apply Transformers in 3D point cloud representation learning \cite{zhao2021point,guo2021pct,yu2021pointr}. Zhao et al. \cite{zhao2021point} propose to use vectorized self-attention mechanism in Transformer-layer, and apply a hierarchical structure with local feature aggregation. Guo et al. \cite{guo2021pct} propose to use neighbour embedding to enhance the representation learning ability. However, previous Transformer-based methods on point cloud representation learning bring in inevitable inductive biases and manual assumptions, the standard Transformer with no inductive bias is proved to perform poorly \cite{yu2021pointbert,han2022dual} due to the limited scale of point cloud data. In this work, we aim to extend the success of standard Transformer to 3D point cloud representation learning.

\begin{figure*}[tb]
  \centering
  \includegraphics[width=2\columnwidth]{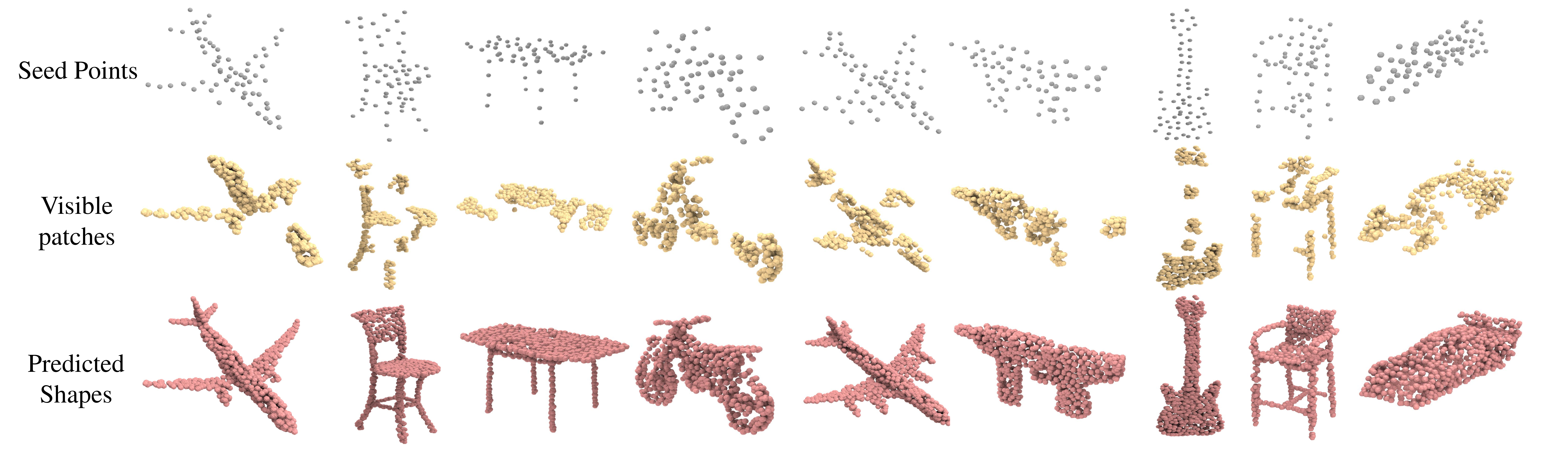}
  \caption{\textbf{Visualization of reconstructing shapes from seed points and visible patches.} We show the reconstructed objects from different categories in the ShapeNet dataset, even including some very complex objects (e.g. Motorcycle). Our model is able to reconstruct the complete shape from a highly sparse seed point cloud which has only 64 points and a highly occluded (only 25$\%$ visible) input point cloud.
  }
  \label{fig:vis_pmae}
\end{figure*}

\subsubsection{Self-supervised Learning on Point Clouds}
Self-supervised learning (SSL) is to learn the representation from unlabelled data, where the supervision signals are built from the data itself. Since the annotation of point clouds is often time-consuming and error-prone, the performance of supervised approaches is difficult to be further improved. Therefore, SSL on point clouds becomes more and more important. Recently, several works were proposed to use SSL techniques for point cloud representation learning\cite{yang2018foldingnet,xie2020pointcontrast,wang2021unsupervised,sun2021point,sauder2019self,sanghi2020info3d,rao2020global,li2018so,eckart2021self,zhang2021self}. Sauder et al. \cite{sauder2019self} propose to rearrange shape parts and reconstruct the original shapes. PointContrast \cite{xie2020pointcontrast} designs a SSL scheme by contrastive learning on different views of point clouds. 
DepthContrast \cite{zhang2021self} was proposed to conduct contrastive learning information from the depth scans. 
CrossPoint \cite{afham2022crosspoint} introduces a cross-modality contrastive learning strategy by exploring self-supervised signals from the semantic differences between point clouds and their rendering images. 
Inspired by BERT, Point-BERT \cite{yu2021pointbert} achieves great performance by pre-training a standard Transformer in a BERT-style SSL scheme. However, Point-BERT only focuses on distinguishing tokens of different patches, makes it difficult for them to transfer to downstream generative tasks.

\emph{\textbf{Auto-encoder.}}  An auto-encoder architecture consists of two parts: an encoder and a decoder. A number of approaches \cite{wang2021unsupervised,yang2018foldingnet,eckart2021self,han2019multi,liu2019l2g} apply auto-encoder architecture to learn meaningful representations from unlabelled point clouds. The propose of point cloud auto-encoder is to learn the presentation from the input shape and then reconstruct the shape from the learned low-dimension latent code. FoldingNet \cite{yang2018foldingnet} designs a point cloud anto-encoding with a folding-based decoder. IAE \cite{yan2022implicit} proposes to use an implicit auto-encoder to learn more detailed geometric information from the additional supervision of the occupancy values and the signed distance values, which prevents IAE from self-supervised learning directly using the point clouds. OcCo \cite{wang2021unsupervised} proposes to complete view-occluded point cloud with a standard point cloud completion network. 
However, these methods only focus on the generation ability of the whole shape, thus mixing the local and global geometry features together, makes it hard to transfer the knowledge to downstream tasks. 
Recently, in 2D vision, He et al. \cite{he2021masked} propose a new form of auto-encoders named MAE by masking regular patches of images and learning to recover the masked parts. Partly inspired by MAE, we design a new self-supervised learning framework to recover the complete shapes from the highly occluded shapes.

\begin{figure*}[tb]
  \centering
  \includegraphics[width=2\columnwidth]{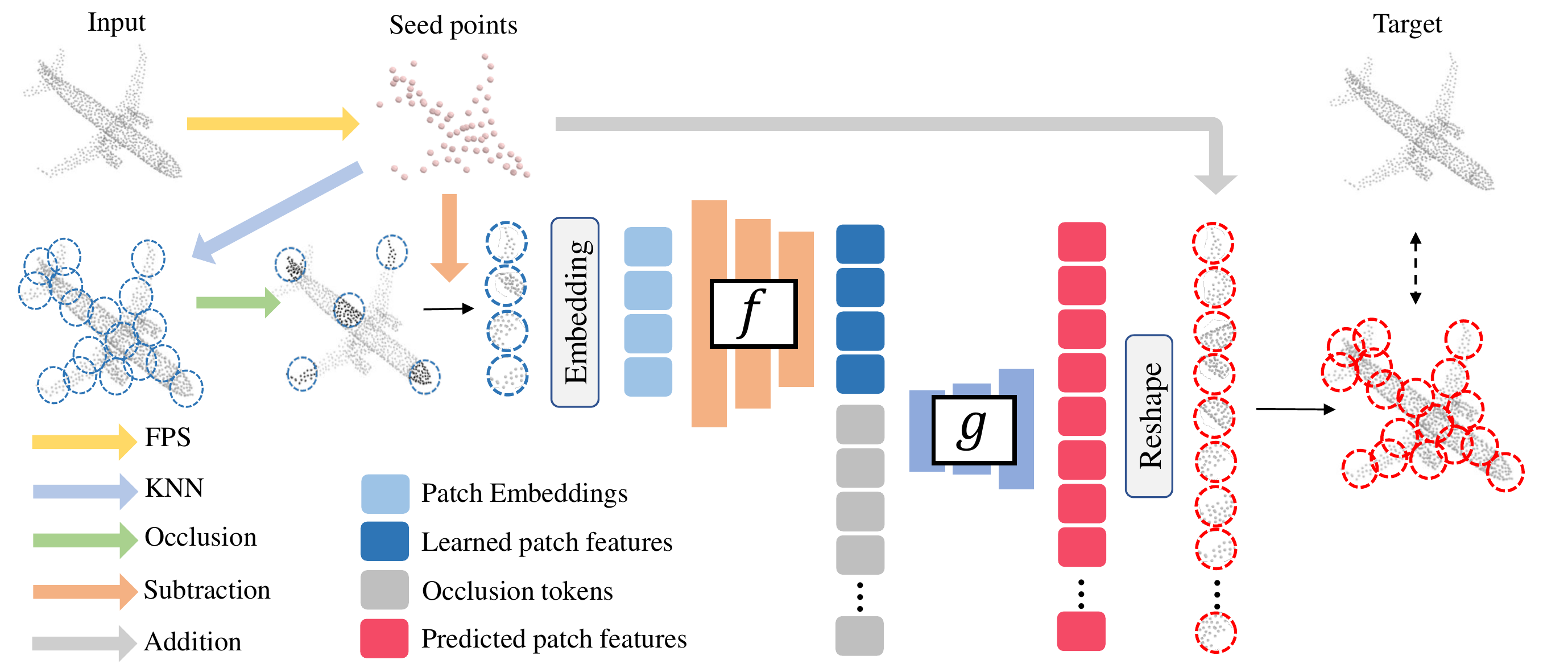}
  \caption{\textbf{Overview of 3D-OAE.} The $f$ and $g$ indicate the standard Transformer-based encoder and decoder. We first extract seed points from the input point cloud using FPS, and then separate the input into point patches by grouping local points around each seed point using KNN. After that, we randomly occlude high ratio of patches and subtract each visible patch to its corresponding seed point for detaching the patch from its spatial location. The encoder $f$ operates only on the embeddings of visible patches and the learnable occlusion tokens are combined to the latent feature before the decoder $g$. Finally, we operate addition to the output patches and their corresponding seed points to regain their spatial locations and further merge the local patches into a complete shape, where we compute a loss function with the ground truth.
  }
  \label{fig:overview}
\end{figure*}

\section{Occlusion Auto-encoder}
The overall architecture of 3D-OAE is shown in Fig. \ref{fig:overview}. Like other point cloud auto-encoders, 3D-OAE consists of an encoder which learns the representation from the input shape and a decoder to reconstruct the original shape from the learned representation. Unlike other point cloud auto-encoders which operates on the whole shape, 
3D-OAE divides the complete shape into groups of patches, highly occludes them, and learns to recover the missing patches of shapes. To achieve this, an asymmetrical encoder-decoder architecture is designed with an encoder only operates 
on the visible subset of patches, and a decoder to predict local features of occluded patches from the visible ones. After that, we combine the predicted local features of occluded patches and their corresponding seed points which serve as global hints to infer the missing geometries that semantically match the input 3D shape. After self-supervised learning, we can leverage the encoder in different downstream tasks as illustrated in Fig. \ref{fig:3D-OAE}. Specifically, we first operate average pooling to aggregate all local features extracted from the trained encoder into a global feature for representing the whole shape, and then fed it into the special decoders of different downstream tasks.

\subsection{Grouping and Occluding}
Previous Transformer-based methods treat each single point in the original shape as a minimum operation unit like words in sentences. However, it brings huge computational complexity and large demand for memory due to the large scale of point cloud data (we don't expect a sentence to have thousands of words). Inspired by previous works \cite{dosovitskiy2020image,yu2021pointbert}, we choose to use patches of point clouds as the minimum unit. To achieve this, we first use Furthest Point Sampling (FPS) to sample seed points $s \in \mathbb{R}^{G\times 3} $ on a given input point cloud $p \in \mathbb{R}^{N\times 3} $, and then use K Nearest-Neighbour (KNN) to sample sets of point patches $\{ g_i | g_i \in \mathbb{R}^{K\times 3} \}$ around each seed point $\{ s_{i} \}^{G}_{i=1}$, as shown in Fig. \ref{fig:overview}.
But it doesn't work to put these patches directly into a neural network because the structure information and spatial coordinates are entangled in point clouds.
We solve this problem by centralizing each patch to its corresponding seed point, thus each patch only contains its local geometry details while seed points provide the global hints.

We apply a straightforward occluding strategy: we randomly select a subset of seed points $\{ s_i \}^{R}_{i=1}$, and then remove their corresponding patches $\{ g_i \}^{R}_{i=1}$. After that, we project each of the remain visible patches $\{ g_i \}^{G-R}_{i=1}$ into a patch embedding as shown in Fig. \ref{fig:overview} using a simple PointNet as:
\begin{equation}
    \label{eq:embed}
    {E_{i}} = Max(x_i) 
    \in \mathbb{R}^{1\times C}, where\  \ 
    {x_{i}} = \phi({g_i}|\theta) \in \mathbb{R}^{K\times C},
\end{equation}
where $\phi$ and $\theta$ denotes the MLP layers and the weights, $C$ is the channel of patch embeddings and $Max$ denotes Max-Pooling operation. The patch embeddings $\{ E_i \}^{G-R}_{i=1}$ will serve as the inputs to the encoder $f$. 

We choose to occlude a very large regions ($75\%$) of the original shape, more numerical comparison of occlusion ratios can be found in Table \ref{table:ablation2}. Removing a high ratio of patches largely increases the difficulty of auto-encoding reconstruction, thus forces the model to learn a powerful representation to generate more detailed local geometries. More importantly, the design of highly occlusion strategy makes it possible for efficient self-supervised learning on large scale unlabelled point cloud data.

\subsection{Transformer}
We will simply review the standard transformer block \cite{vaswani2017attention} which serves as the unit architecture of our proposed 3D-OAE in this section. A transformer block consists of a self-attention layer with multi-heads and a feed-forward network which is implemented as a single hidden layer. Given a set of patch embeddings $\{ E_i \}^{G-R}_{i=1}$ as the input, we first map them into queries, keys and values which are formulated as matrices $Q$, $K$, and $V$ using MLPs. The dot-product self-attention is then computed by:
\begin{equation}
    A(Q, K, V)=softmax(\frac{QK^T}{\sqrt{d}}V),
\end{equation}
where $d$ means the channel dimension. A parallel multi-head strategy is further applied for allowing the transformer to attend to the information from different subspace jointly,
\begin{equation}
\begin{split}
    M(Q, K, V)= 
     Concat&(a_1, a_2, ..., a_h)W,
    \\ where \ &a_i = A(QW_i^Q, KW_i^k, VW_i^V)
\end{split}
\end{equation}
and $W, W_i^Q, W_i^k, W_i^V$ are projection matrices with learnable parameters.

\subsection{Auto-encoder Architecture}

\subsubsection{3D-OAE Encoder}
We adopt the 3D point cloud standard Transformers with multi-headed self-attention layers and FFN blocks as detailed above as the unified backbone of our architecture. Specifically, our encoder is a standard Transformer but applies only on visible patches. 
For the input visible patches, we first extract their patch embeddings as described in Eq. (\ref{eq:embed}). To distinguish centralized patches apart, we use a simple MLP $\gamma$ to extract the position embeddings of visible seed points $\{ s_i \}^{G-R}_{i=1}$ and add them to their corresponding patch embeddings as:
\begin{equation}
    {\mathbb{E}_{i}} \gets {\gamma(s_i|\theta)} + {E_{i}},
\end{equation}
After that, a series of Transformer blocks is applied to these patch embeddings to learn representations. 
\begin{equation}
     {\mathbb{E^{'}}} = Linear(f_\varphi(\mathbb{E},H_e)),
\end{equation}
where $f_\theta$ indicates the Transformer encoder, $H_e$ represents the number of Transformer blocks in $f_\theta$, and $\mathbb{E}=[\mathbb{E}_{1}, \mathbb{E}_{2}, ..., \mathbb{E}_{G-R}]$ is the set of patch embeddings. A linear projection layer is further applied for dimension mapping.

Since we use a very high occlusion ratio, the encoder operates only on a small subset (e.g. 25$\%$) of patches, which makes it possible to do self-supervised learning in very large scale unlabelled data with a relatively huge encoder.

\subsubsection{3D-OAE Decoder}
The input of 3D-OAE decoder is a full set of patch embeddings consisting of the encoded visible patch embeddings and the occlusion tokens $\{ T_i \}^{R}_{i=1}$, fomulated as:
\begin{equation}
     {\mathbb{U}} = Concat(\mathbb{E^{'}}, T),
\end{equation}
where $\mathbb{E^{'}}\in \mathbb{R}^{(G-R)\times K\times C}$, $T\in \mathbb{R}^{R\times K\times C}$ and $\mathbb{U}\in \mathbb{R}^{G\times K\times C}$.

Each occlusion token is a shared, learnable vector which aim at learning to reconstruct one occluded patch. We further add the position embeddings $\{\gamma(s_i|\theta)\}^{G}_{i=1}$ to the full set of patch embeddings for providing the location information to the occlusion tokens.
Then a series of light-weighted Transformer blocks are further applied to learn the occlusion tokens from features of visible 
patches via self-attention mechanism:
\begin{equation}
     {\mathbb{U^{'}}} = f_\omega(\mathbb{U}+\{\gamma(s|\theta)\},H_d),
\end{equation}
where $f_\omega$ and $H_d$ are the Transformer decoder and the number of blocks of it.

Since we calculate the attention map of each patch embedding to all of the others, the model will have no sensitivity about the ordering of patches, which indicates that 3D-OAE is suitable for the unodered point cloud data. 

The decoder $g$ is only used during self-supervised learning to recover the occluded parts of the original shape, only the learned encoder $f$ is used when transferring to downstream tasks (e.g. object classification, point cloud segmentation, point cloud completion), which means we don't care much about the learning ability of the decoder. Therefore, we design a light-weighted decoder with only about 20$\%$ computation of the encoder. And the training process is largely accelerated since the full set of patch embeddings is only processed by the light-weighted decoder.

\subsection{Optimization Objective}
During training, the goal of 3D-OAE is to reconstruct the complete shape from seed points and visible point patches. After encoding and decoding, 3D-OAD outputs patch-wised vectors where each vector contains the local geometry information of a single patch. The feature channel of the Transformer decoder $f_\omega$ is set to be the product of point dimensions and patch point numbers, thus each vector can be directly reshaped to the size of a local patch. Finally, the seed points are added to their corresponding patches to reconstruct the complete shape. We choose Chamfer Distance described by Eq. (\ref{eq:cd}) as our loss function. Similar to BERT \cite{devlin2018bert} and MAE \cite{he2021masked}, we don't pay much attention to the reconstruction ability of visible parts, and only compute loss between the points of predicted patches $\mathcal{P}^{\rm o}$ and the ground truth point cloud of these occluded patches denoted as $\mathcal{P}^{\rm t}$. 

\begin{equation}\label{eq:cd}\small
\begin{split}
  \mathcal{L}_{\rm CD}(\mathcal{P}^{\rm o},\mathcal{P}^{\rm t})=&{\frac{1}{|\mathcal{P}^{\rm {\rm o}}|}}\sum_{\bs{p}^{\rm o}\in \mathcal{P}^{\rm {\rm o}}}\min_{\bs{p}^{\rm t}\in \mathcal{P}^{\rm t}} {\|\bs{p}^{\rm o}-\bs{p}^{\rm t} \|}_2 
  \\&+{\frac{1}{|\mathcal{P}^{\rm {\rm t}}|}} \sum_{\bs{p}^{\rm t}\in \mathcal{P}^{\rm t}}\min_{\bs{p}^{\rm o}\in \mathcal{P}^{\rm o}}{\|\bs{p}^{\rm t}-\bs{p}^{\rm o} \|}_2.
\end{split}
\end{equation}

We also try to use Earth Mover's Distance as the loss function but find it unhelpful, please see the numerical comparison in Table \ref{table:ablation1}.





\section{Experiments and Applications}

In this section, we first introduce the self-supervised learning setting of 3D-OAE in Sec. \ref{section:sec4.1}. Next, we evaluate the proposed model with various downstream tasks in Sec. \ref{section:sec4.2} - Sec. \ref{section:sec4.6}, including shape understanding, few-shot learning, part segmentation and transfer learning to generative tasks and real world tasks. We show the efficiency and learning curves in \ref{section:efficiency} and \ref{section:curves}. Finally, We conduct ablation studies on framework designs and occlusion ratios in Sec. \ref{section:sec4.7}.

\subsection{Self-supervised Learning}
\label{section:sec4.1}
\noindent\textbf{Dataset.} We learn the self-supervised representation model from the ShapeNet\cite{chang2015shapenet} dataset  which contains 57,448 synthetic models from 55 categories.
We sample 1024 points from each 3D model and divide them into 64 point cloud patches using Furthest Point Sample (FPS) and K-Nearest Neighbor (KNN), where each patch contains 32 points. During training, we apply the same data augmentations as PointNet++ \cite{qi2017pointnet++}.

\noindent\textbf{Training setups.} In the self-supervised learning stage, we set the Transformer depth of both encoder and decoder to 12 and the number of Transformer heads both to 6. The feature channel dimension of encoder and decoder Transformers are set to 384 and 96, 
and the occlusion ratio is set to 75${\%}$.
We adopt an AdamW \cite{loshchilov2017decoupled} optimizer, using an initial learning rate of 0.0005 and a weight decay of 0.05. And we train our model for 300 epochs with a batch size of 256 on one 2080Ti. 

\begin{figure}[!t]
  \centering
  \includegraphics[width=\columnwidth]{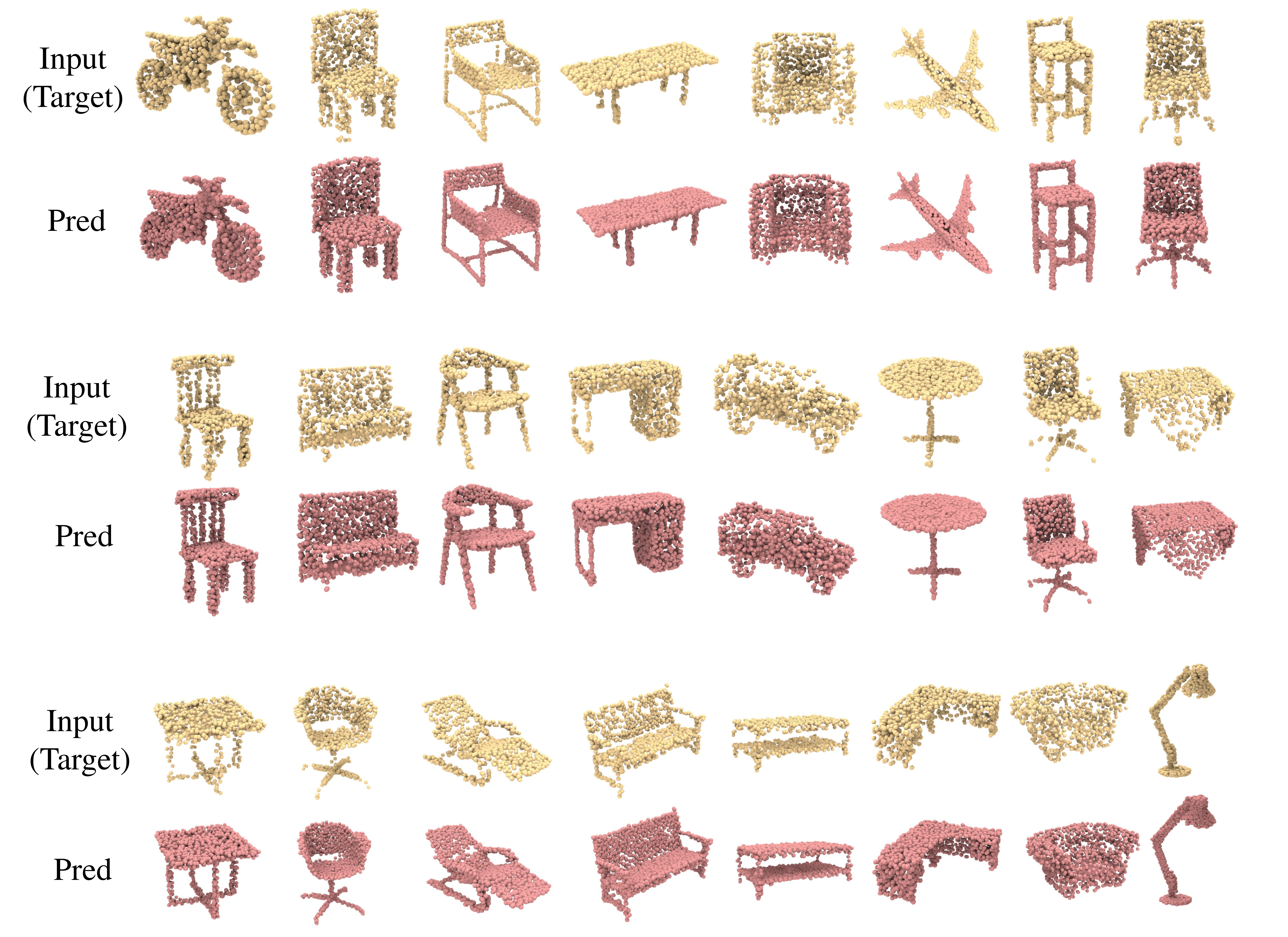}\vspace{-0.2cm}
  \caption{\textbf{Visualization of self-reconstruction results. }We show the auto-encoder results of our 3D-OAE, where the first row is the input shapes which also indicate the target shapes, and the second row is the reconstruction results of 3D-OAE. Both the input shapes and the predicted shapes contain 1024 points. }
  \label{fig:support_ae}
\end{figure}

\noindent\textbf{Visuliazation. }In Fig. \ref{fig:support_ae}, we present more self-reconstruction results of 3D-OAE as a supplement of Fig.\ref{fig:vis_pmae}. We provide the visualization of both the reconstructed shapes and the complete input shapes which also indicate the target shapes. We show the complete input shapes for visual comparison but still notice that 3D-OAE only operates on a small subset patches of the input shapes. 

\subsection{Shape Understanding}
\label{section:sec4.2}
We follow prior works \cite{yang2018foldingnet,han2019multi,zhao20193d} to evaluate the shape understanding capability of our self-supervision model using the ModelNet40 \cite{wu20153d} benchmark. 
It contains 12,311 synthesized models from 40 categories and is split into 9843/2468 models for training and testing. 
After training our models in ShapeNet \cite{chang2015shapenet} as detailed in Sec. \ref{section:sec4.1}, we evaluate the learned representation of trained encoder in the following experiments.

\setlength{\tabcolsep}{5mm}
\begin{table}
\begin{center}
\caption{{\bf{Linear evaluation for shape classification on ModelNet40.}} A linear classifier is trained on the representation learned from the ShapeNet dataset by different self-supervision methods. STransformer means 3D standard Transformer.}
\label{table:linearsvm}
\begin{tabular}{lcc}
\hline\noalign{\smallskip}
Method & Input &  Accuracy\\
\noalign{\smallskip}
\hline
\noalign{\smallskip}
3D-GAN\cite{wu2016learning} & voxel & 83.3${\%}$\\
VIP-GAN\cite{han2019view} & views & 90.2${\%}$\\
Latent-GAN\cite{valsesia2020learning} & points & 85.7${\%}$\\
SO-Net\cite{li2018so} & points & 87.3${\%}$\\
FoldingNet\cite{yang2018foldingnet} & points & 88.4${\%}$\\
MRTNet\cite{gadelha2018multiresolution} & points & 86.4${\%}$\\
3D-PointCapsNet\cite{zhao20193d} & points & 88.9${\%}$\\
MAP-VAE\cite{han2019multi} & points & 88.4${\%}$\\
PointNet + Jiasaw\cite{sauder2019self} & points & 87.3${\%}$\\
DGCNN + Jiasaw\cite{sauder2019self} & points & 90.6${\%}$\\
PointNet + Orientation\cite{poursaeed2020self} & points & 88.6${\%}$\\
DGCNN + Orientation\cite{poursaeed2020self} & points & 90.7${\%}$\\
PointNet + STRL\cite{huang2021spatio} & points & 88.3${\%}$\\
DGCNN + STRL\cite{huang2021spatio} & points & 90.9${\%}$\\
PointNet + CrossPoint\cite{afham2022crosspoint} & points & 89.1${\%}$\\
DGCNN + CrossPoint\cite{afham2022crosspoint} & points & 91.2${\%}$\\
\hline
STransformer + OcCo\cite{wang2021unsupervised} & points & 89.6${\%}$\\
STransformer + Point-BERT\cite{yu2021pointbert} & points & 87.4${\%}$\\
3D-OAE (Ours) & points & \bf{92.3${\%}$}\\
\hline
\end{tabular}
\end{center}
\end{table}

\subsubsection{Linear SVM}
In this experiment, we train a linear Support Vector Machine (SVM) classifier using the representation from our trained encoder Transformer. The number of point clouds is down-sampled to 2048 for both training and testing. The comparison of classification results is shown in Table \ref{table:linearsvm}. Our proposed 3D-OAE achieves state-of-the-art performance of 92.3${\%}$ accuracy on test sets, while the runner-up method only achieves 91.2${\%}$ accuracy. It's worth noting that this result has reached the accuracy of training a classification network from scratch (e.g. PointNet++ (90.5${\%}$), DGCNN (92.2${\%}$)), which proves that the representation domain learned by 3D-OAE is highly decoupled. 
Since our model is learned on the ShapeNet dataset, we believe that this result also shows the strong transfer ability of our model.

\subsubsection{Supervised Fine-tuning}
In this experiment, we explore the ability of our model to transfer to downstream classification tasks. The {\bf{supervised}} models are trained from scratch and the {\bf{self-supervised}} models use the trained weights from self-supervised learning as the initial weights for fine-tuning. All the self-supervised methods use the standard Transformer (STransformer) as backbone architecture. 
For a fair comparison with OcCo, we follow the details illustrated in \cite{wang2021unsupervised} and use the standard Transformer encoder and Transformer-based decoder PoinTr \cite{yu2021pointr} to reproduce the completion task in ShapeNet. 
In comparison, our 3D-OAE brings 2.0${\%}$(91.4${\%}$/93.4${\%}$) accuracy improvement over training from scratch. And our method also outperforms PCT \cite{guo2021pct}, which is a variety of standard Transformer. The result proves that using our self-supervised learning scheme, a standard Transformer with no inductive bias could also learn a powerful representation.

\setlength{\tabcolsep}{4pt}
\begin{table}
\begin{center}
\caption{{\bf{Shape classification results fine-tuned on ModelNet40}}}
\label{table:finetune}
\begin{tabular}{c|lc}
\hline\noalign{\smallskip}
Category & Method &  Accuracy\\
\noalign{\smallskip}
\hline
\noalign{\smallskip}
\multirow{6}*{Supervised} & PointNet\cite{qi2017pointnet} & 89.2${\%}$\\
~ & PointNet++\cite{qi2017pointnet++} & 90.5${\%}$\\
~ & PointCNN\cite{li2018pointcnn} & 92.2${\%}$\\
~ & DGCNN\cite{wang2019dynamic} & 92.2${\%}$\\
~ & PCT\cite{guo2021pct} & 93.2${\%}$\\
~ & STransformer & 91.4${\%}$\\
\hline
\multirow{3}*{Self-supervised} & STransformer + OcCo\cite{wang2021unsupervised} & 92.1${\%}$\\
~ & STransformer + Point-BERT\cite{yu2021pointbert} & 93.2${\%}$\\
~ & 3D-OAE (Ours) & \bf{93.4}${\%}$\\
\hline
\end{tabular}
\end{center}
\end{table}
\setlength{\tabcolsep}{1.4pt}

\begin{figure*}[!t]
  \centering
  \includegraphics[width=0.95\textwidth]{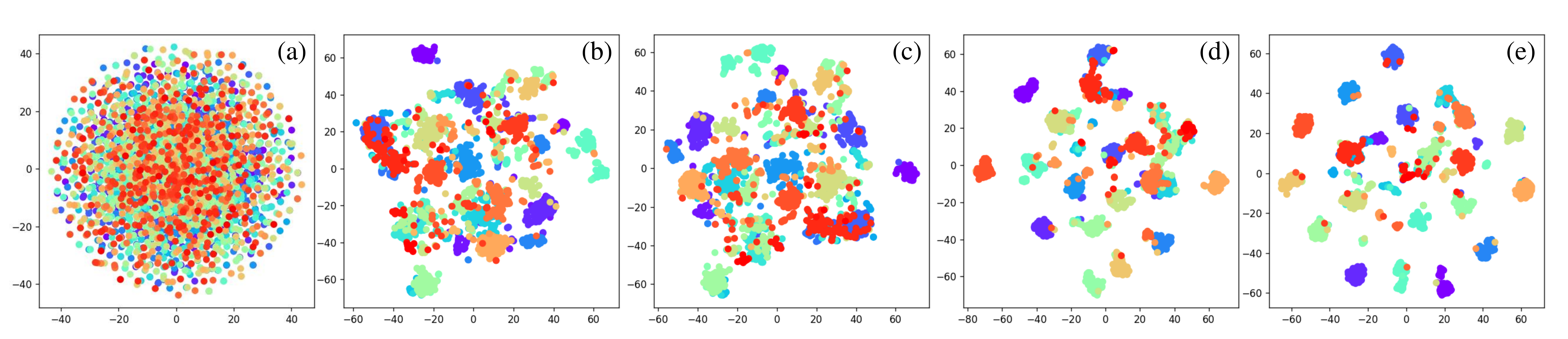}
  \caption{\textbf{Visualization of feature distributions.} We visualize the features of test sets in ModelNet40 using t-SNE. (a) random initialization, (b) 3D-OAE pre-trained on ShapeNet, (c) Point-BERT pre-trained on ShapeNet, (d) train an randomly initialized encoder on ModelNet40,  (e) fine-funing learned encoder of 3D-OAE on ModelNet40.
  }
  \label{fig:t-sne}
\end{figure*}

\subsubsection{Embedding Visualizations}
We visualize the feature distributions using t-SNE \cite{van2008visualizing}. Fig. \ref{fig:t-sne} (b) shows the features learned by 3D-OAE after self-supervised training on the ShapeNet dataset.
It's clear that the feature space of different categories which are mixed together in random initialization (Fig. \ref{fig:t-sne} (a)) can be well separated into different regions by 3D-OAE.  We also achieve comparable performance with Point-BERT(Fig. \ref{fig:t-sne} (c)).
As shown in Fig. \ref{fig:t-sne} (e), the feature space are almost separated completely independent after fine-funing on ModelNet40 train sets, and are more clearly disentangled than training from scratch (Fig. \ref{fig:t-sne} (d)). It proves that 3D-OAE can guide Transformers to learn powerful representations from unlabelled data, even from different dataset.

\setlength{\tabcolsep}{6pt}
\begin{table*}
\begin{center}
\caption{{\bf{Few-shot classification results on ModelNet40}}}
\label{table:fewshot}
\resizebox{12cm}{!}{
\begin{tabular}{lccccc}
\noalign{\smallskip}
\hline
\noalign{\smallskip}
\multirow{2}*{} & \multicolumn{2}{c}{5 way} & & \multicolumn{2}{c}{10 way}\\
\cline{2-3}
\cline{5-6}
~ & 10-shot & 20-shot & & 10-shot & 20-shot\\
\hline
DGCNN-rand \cite{wang2019dynamic} & 91.8 $\pm$ 3.7 & 93.4 $\pm$ 3.2 & & 86.3  $\pm$ 6.2 & 90.9  $\pm$ 5.1 \\
DGCNN-OcCo \cite{wang2021unsupervised} & 91.9  $\pm$ 3.3 & 93.9 $\pm$ 3.1 & & 86.4  $\pm$ 5.4 & 91.3  $\pm$ 4.6 \\
STransformer-rand & 87.8  $\pm$ 5.2 & 93.3 $\pm$ 4.3 & & 84.6  $\pm$ 5.5 & 89.4  $\pm$ 6.3 \\
STransformer-OcCo \cite{wang2021unsupervised} & 94.0  $\pm$ 3.6 & 95.9 $\pm$ 2.3 & & 89.4  $\pm$ 5.1 & 92.4  $\pm$ 4.6 \\
STransformer-Point-BERT \cite{yu2021pointbert} & 94.6  $\pm$ 3.1 & 96.3 $\pm$ 2.7 & & 91.0  $\pm$ 5.4 & 92.7  $\pm$ 5.1 \\
3D-OAE & \textbf{96.3}  $\pm$ \textbf{2.5} & \textbf{98.2} $\pm$ \textbf{1.5} & & \textbf{92.0}  $\pm$ \textbf{5.3} & \textbf{94.6}  $\pm$ \textbf{3.6} \\
\hline
\end{tabular}}
\end{center}
\end{table*}
\setlength{\tabcolsep}{1.4pt}

\subsection{Few-shot Learning}
\label{section:sec4.3}
We further evaluate our model by conducting few-shot learning experiments on ModelNet40. A common used setting is ``K-way N-shot", where K classes are first random selected, and then (N+20) samples are sampled from each class. The model is trained on K ${\times}$ N samples, and evaluated on K ${\times}$ 20 samples. Following previous work \cite{sharma2020self,yu2021pointbert}, we choose 4 different few-shot learning settings: ``5 way, 10 shot", ``5 way, 20 shot", ``10 way, 10 shot" and ``10 way, 20 shot". For fair comparison, we use the data processed by Point-BERT \cite{yu2021pointbert} to conduct 10 separate experiments on each few-shot setting. Table \ref{table:fewshot} reports the mean accuracy and standard deviation of these 10 runs. 

We compare our model with currently state-of-the-art methods OcCo \cite{wang2021unsupervised} and Point-BERT \cite{yu2021pointbert}. As shown in Table \ref{table:fewshot}, using standard Transformer as backbone, our proposed 3D-OAE achieves a significant improvement of 8.5${\%}$, 4.9${\%}$, 7.4${\%}$, 5.2${\%}$ over baseline, and 1.7${\%}$, 1.9${\%}$, 1.0${\%}$, 1.9${\%}$ over the runner-up method Point-BERT in 4 different sets of few-shot classification. 
We even achieve 98.2${\%}$ accuracy on the ``5 way 20 shot" with a standard deviation of only 1.5. 
The outstanding performance on few-shot learning proves the strong ability of 3D-OAE to transfer to downstream tasks using very limited data. 

\subsection{Object Part Segmentation}
\label{section:sec4.4}
Object part segmentation is a challenging task which aims to predict the part label for each point of the model. The ShapeNetPart \cite{yi2016scalable} dataset consists of 16,800 models from 16 categories and is split into 14006/2874 for training and testing. 
The number of parts for each category is between 2 and 6, and there are 50 different parts in total.
We sample 2048 points from each model follow PointNet \cite{qi2017pointnet}, and apply a segmentation head achieved by Point-BERT \cite{yu2021pointbert} to propagates the group features to each point hierarchically. 

\begin{figure*}[!t]
  \centering
  \includegraphics[width=2\columnwidth]{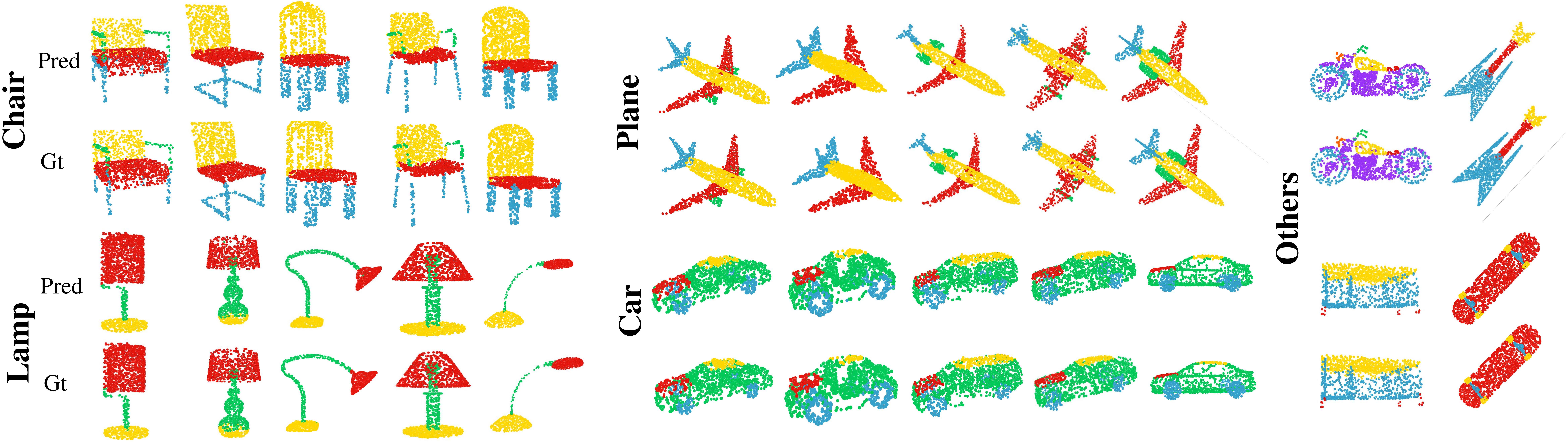}
  \caption{\textbf{Visualization of our part segmentation results. } Different colors indicate different parts. Top row shows results predicted by our model; bottom row shows the corresponding ground truth.
  }
  \label{fig:vis_seg}
\end{figure*}

As shown in Table \ref{table:part_seg}, our model achieves 0.6$\%$ improvement over training a standard Transformer from scratch, while OcCo fails to improve performance. 3D-OAE also outperforms PointNet, PointNet++ and DGCNN. A visualization of out segmentation results is shown in Fig. \ref{fig:vis_seg}. It's clear that our model is able to make the right predictions at most points, and there are only small visual differences between our predictions and the ground truth.

\begin{table*}[!t]\small
\setlength{\tabcolsep}{1mm}
\centering
\caption{\textbf{Part segmentation results on the ShapeNetPart dataset.} We report the mean IoU across all instance and IoU for each categories.}
\resizebox{17cm}{!}{
\begin{tabular}{l|c|cccccccccccccccc}
\toprule
Methods &mIoU$_I$  &aero    &bag  &cap   &car   &chair   &ear.    &guitar    &knefe & lamp & lap. & motor & mug & pistol & rock. & skate. & table  \\ 
\midrule
PointNet\cite{qi2017pointnet} & 83.7 & 83.4 & 78.7 &  82.5 & 74.9 & 89.6 & 73.0 & 91.5 & 85.9 & 80.8 & 95.3 & 65.2 & 93 & 81.2 & 57.9 & 72.8 & 80.6 \\
PointNet++\cite{qi2017pointnet++} & 85.1 & 82.4 & 79 & 87.7 & 77.3 & 90.8 & 71.8 & 91 & 85.9 & 83.7 & 95.3 & 71.6 & 94.1 & 81.3 & 58.7 & 76.4 & 82.6\\
DGCNN\cite{wang2019dynamic} & 85.2 & 84 & 83.4 & 86.7 & 77.8 & 90.6 & 74.7 & 91.2 & 87.5 & 82.8 & 95.7 & 66.3 & 94.9 & 81.1 & 63.5 & 74.5 & 82.6\\
\midrule
STransformer  & 85.1 & 82.9 & 85.4 & 87.7 & 78.8 & 90.5 & 80.8 & 91.1 & 87.7 & 85.3 & 95.6 & 73.9 & 94.9 & 83.5 & 61.2 & 74.9 & 80.6\\
STransformer-OcCo \cite{wang2021unsupervised}   & 85.1 & 83.3 & 85.2 & 88.3 & 79.9 & 90.7 & 74.1 & 91.9 & 87.6 & 84.7 & 95.4 & 75.5 & 94.4 & 84.1 & 63.1 & 75.7 & 80.8\\
STransformer-Point-Bert \cite{yu2021pointbert} & 85.6 & 84.3 & 84.8 & 88.0 & 79.8 & 91.0 & 81.7 & 91.6 & 87.9 & 85.2 & 95.6 & 75.6 & 94.7 & 84.3 & 63.4 & 76.3 & 81.5\\
3D-OAE	&\textbf{85.7}  & 83.4 & 85.0 & 83.8 & 79.3 & 80.1 & 80.1 & 91.9 & 87.2 & 82.5 & 95.3 & 76.0 & 95.1 & 85.6 & 63.5 & 80.5 & 83.6\\
\bottomrule
\end{tabular}}
\label{table:part_seg}
\end{table*}

\subsection{Transfer to Generative Tasks}
\label{section:sec4.5}
Since most of the previous self-supervised learning methods only focus on the discriminant ability of the representation learned by their model and verify it by transferring the model to classification tasks. They fail to transfer their model to downstream generative tasks (e.g. point cloud completion, point cloud up-sampling). In this section, we show the transfer learning ability of 3D-OAE to downstream generative tasks by conducting point cloud completion experiments.

\noindent\textbf{Dataset briefs and evaluation metric.}
The PCN \cite{yuan2018pcn} dataset is one of the most widely used benchmark datasets in point cloud completion task. 
It contains 30,974 models from 8 categories. For each 3D object, 16,384 points are sampled from the shape surface, and 8 partial point clouds are generated by back-projecting 2.5D depth images from 8 views into 3D. 
For a fair comparison, we use the same train/test split settings of PCN\cite{yuan2018pcn} and follow previous works to adopt the L1 version of Chamfer distance for evaluation. 
We use a standard Transformer encoder and a Transformer-based decoder proposed in PoinTr \cite{yu2021pointr} as our backbone, and OcCo is trained using the same architecture. We finetune our model and OcCo under PCN dataset on a single GTX 3090Ti GPU, and it takes 300 epochs to converge.

\begin{table*}[!t]\small
\centering
\setlength{\tabcolsep}{2mm}
\caption{\textbf{Point cloud completion on PCN dataset.} The results is reported in terms of per-point L1 Chamfer distance $\times 10^{3}$ (lower is better).}
\resizebox{14.5cm}{!}{
\begin{tabular}{l|c|cccccccc}
\toprule
Methods &Average  &Plane    &Cabinet  &Car   &Chair   &Lamp   &Couch    &Table    &Boat   \\ 
\midrule
FoldingNet  \cite{yang2018foldingnet}   &14.31  &9.49    &15.80    &12.61    &15.55   &16.41    &15.97    &13.65    &14.99   \\
TopNet  \cite{tchapmi2019topnet}     &12.15   &7.61   &13.31   &10.90    &13.82    &14.44      &14.78   &11.22  &11.12   \\
AtlasNet \cite{groueix2018papier}   &10.85   &6.37    &11.94    &10.10    &12.06    &12.37    &12.99    &10.33    &10.61   \\
PCN  \cite{yuan2018pcn}   &9.64 &5.50    &22.70    &10.63    &8.70    &11.00    &11.34    &11.68    &8.59   \\
GRNet \cite{xie2020grnet}  &8.83   &6.45   &10.37   &9.45    &9.41    &7.96      &10.51   &8.44  &8.04   \\
CDN \cite{wang2020cascaded}  &8.51   &4.79   &9.97   &8.31    &9.49    &8.94      &10.69   &7.81  &8.05   \\
PMP-Net \cite{wen2021pmp} &8.73  &5.65    &11.24    &9.64    &9.51    &6.95    &10.83    &8.72   &7.25 \\
NSFA \cite{zhang2020detail}  &8.06 &4.76 &10.18 &8.63 &8.53 &7.03 &10.53 &7.35 &7.48 \\
PoinTr \cite{yu2021pointr} & 8.38 &4.75 & 10.47 & 8.68 & 9.39 & 7.75 & 10.93 & 7.78 & 7.29\\
SnowflakeNet \cite{xiang2021snowflakenet}	&7.21 &4.29	&9.16	&8.08	&7.89	&6.07	&9.23	&6.55	&6.40 \\
\midrule
STransformer-Scratch    & 7.37 & 4.28 & 9.45 & 8.21 & 7.99 & 6.35 & 9.38 & 6.77 & 6.50\\
STransformer-OcCo\cite{wang2021unsupervised}   & 7.11 & 4.07 & 9.16 & 8.00 & 7.65 & 6.08 & 9.26 & 6.41 & 6.27\\
3D-OAE	&\textbf{6.97} &\textbf{3.99}	&\textbf{8.98}	&\textbf{7.90}	&\textbf{7.46}	&\textbf{5.96}	&\textbf{8.96}	&\textbf{6.31}	&\textbf{6.19} \\
\bottomrule
\end{tabular}}
\label{table:pcn}
\end{table*}

\begin{figure*}[!t]
  \centering
  \includegraphics[width=2\columnwidth]{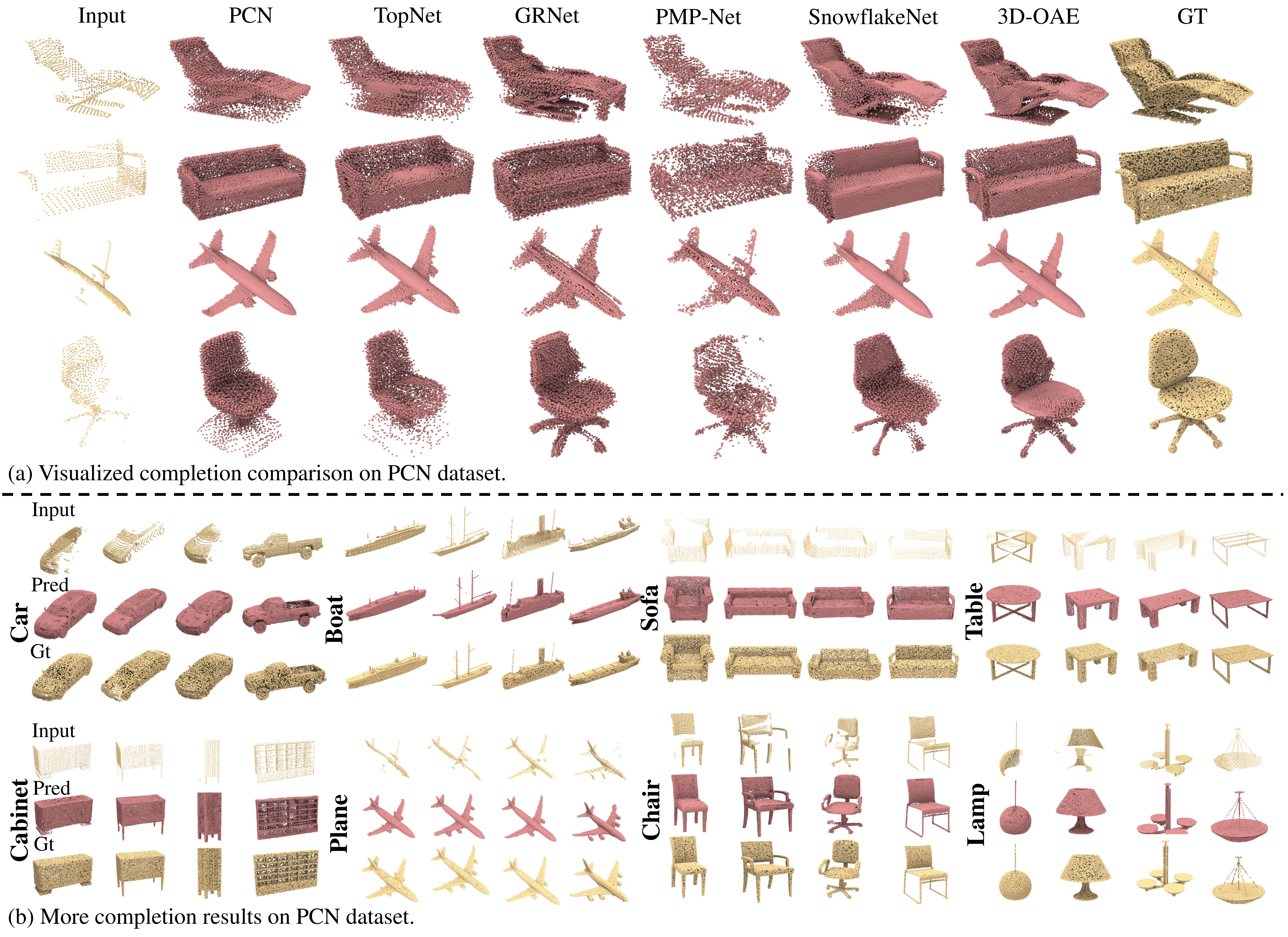}
  \caption{\textbf{Visual comparison of point cloud completion on PCN dataset. } The input and ground truth have 2048 and 16384 points, respectively. We show the completion comparison with current state-of-the-art completion methods like GRNet\cite{xie2020grnet}, PMP-Net\cite{wen2021pmp} and SnowflakeNet\cite{xiang2021snowflakenet} in (a), our 3D-OAE generates more smooth surfaces and more detailed structures. We also provide more completion results of our 3D-OAE in (b).}
  \label{fig:pcn}
\end{figure*}

\noindent\textbf{Quantitative comparison.}
The results of our proposed 3D-OAE and other completion methods are shown on Table \ref{table:pcn}, where 3D-OAE achieves the state-of-the-art performance over all counterparts compared with both supervised and self-supervised methods. We found that 3D-OAE reduces the average CD by 0.4 compared with training a standard Transformer from scratch. 
Especially, 3D-OAE with only a standard Transformer-based model reduces the average CD by 0.24 compared with the state-of-the-art supervised method SnowflakeNet \cite{xiang2021snowflakenet} which proposed a carefully designed model to improve the performance on point cloud completion task. And 3D-OAE also reduces the average CD by 0.14 compared with the state-of-the-art self-supervised method OcCo \cite{wang2021unsupervised} which also takes a generation task to learn the self-supervised representation. These results prove that our generative self-supervised learning framework is able to learn a powerful representation which can bring significant improvement in downstream generative tasks, we may find a way to avoid spending lots of efforts on designing complex and heavy network structures. It is also worth noticing that we only use 1024 points during self-supervised learning, but transfer well to PCN dataset where the input has 2048 points. The visual comparisons of point cloud completion on PCN dataset is shown in Fig. \ref{fig:pcn}.

\begin{table}[!t]\small
\centering
\caption{\textbf{Classification results on the ScanObjectNN dataset.} }
\begin{tabular}{l|ccc}
\toprule
Methods &  OBJ-BG  &OBJ-ONLY   &PB-T50-RS \\ 
\midrule
PointNet\cite{qi2017pointnet} & 73.3 & 79.2 & 68.0 \\
PointNet++\cite{qi2017pointnet++} & 82.3 & 84.3 & 77.9\\
SpiderCNN\cite{xu2018spidercnn} & 77.1 & 79.5 & 73.7\\
PointCNN\cite{li2018pointcnn} & 86.1 & 85.5 & 78.5\\
DGCNN\cite{wang2019dynamic} & 82.8 & 86.2 & 78.1\\
BGA-DGCNN\cite{uy2019revisiting} & - & - & 79.7\\
BGA-PointNet++\cite{uy2019revisiting} & - & - & 80.2\\
\midrule
STransformer & 79.86 & 80.55 & 77.24\\
STransformer-OcCo\cite{wang2021unsupervised} & 84.85 & 85.54 & 78.79\\
STransformer-PBERT\cite{yu2021pointbert} & 87.43 & 88.12 & 83.07\\
3D-OAE & \textbf{89.16} & \textbf{88.64} & \textbf{83.17}\\
\bottomrule
\end{tabular}
\label{table:objnn}
\end{table}

\subsection{Transfer to Real-World Data}
\label{section:sec4.6}
To further evaluate the representation ability of our method, we use the encoder of 3D-OAE which is trained on the synthetic ShapeNet dataset to fine-tune on a real-world dataset ScanObjectNN, which contains 2902 scanned object instances from 15 categories. 
Due to the existence of background, occlusions and noise, this benchmark poses significant challenges to existing methods. We follow previous works to conduct experiments on three main variants: OBJ-BG, OBJ-ONLY and PB-T50-RS. 

As shown in Table \ref{table:objnn}, our proposed 3D-OAE brings significant improvement of 9.03$\%$, 8.09$\%$ , 5.93$\%$ over training a standard Transformer from scratch, and also outperforms the state-of-the-art methods OcCo and Point-BERT. The strong results show that using our self-supervised framework could learn meaningful information from artifical synthetic data and transfer it to real-world data, which could partly solve the domain gap between synthetic and scanned 3D data.

\subsection{Efficiency}
\label{section:efficiency}
\begin{table}[h]\small
\centering
\setlength{\tabcolsep}{2mm}
\caption{\textbf{Efficiency comparison results.} }
\begin{tabular}{l|ccc}
\toprule
Methods &  OcCo \cite{wang2021unsupervised}  &Point-BERT \cite{yu2021pointbert}   &3D-OAE \\ 
\midrule
FLOPs (G) & 8.7 & 9.79 & \textbf{0.65} \\
EpochTime (s) & 1438 & 688 &  \textbf{231}\\
\bottomrule
\end{tabular}
\label{table:efficient}
\end{table}
In Table \ref{table:efficient}, we show the efficiency of our 3D-OAE compared with other point cloud self-supervised learning methods. For a fair comparison, all the methods is trained using a single 2080Ti GPU. The results show that our proposed 3D-OAE has extremely low computational complexity of only 0.65G FLOPs, which is more than 10 times lower than OcCo and Point-BERT. And 3D-OAE also achieves about 6 times faster than OcCo and 3 times faster than Point-BERT. These outstanding results show the efficiency of 3D-OAE, 
due to the designed asymmetric encoder-decoder architecture of 3D-OAE. 
Using our 3D-OAE, it takes only less than one day to train on the full set of ShapeNet dataset for 300 epochs using a single 2080Ti. We can see the possibility of efficient pre-training on large-scale real scanned point cloud data using our framework.

\begin{figure}[h]
\centering
  \includegraphics[width=7cm]{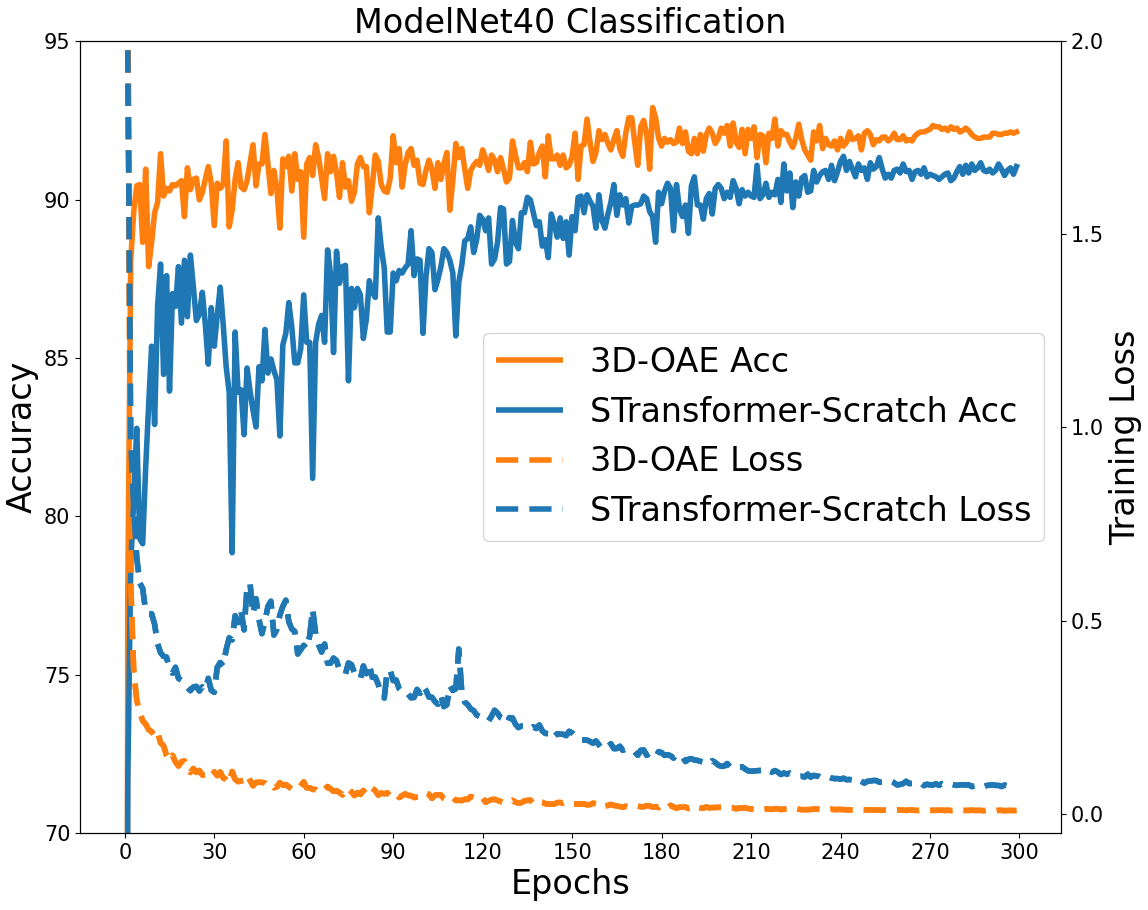}
  \caption{\textbf{Visualization of learning curves in ModelNet40.}}
  \label{fig:mn40}
\end{figure}

\subsection{Learning Curves}
\label{section:curves}
In Fig. \ref{fig:mn40}, we present the visualization of learning curves of both training a standard Transformer from scratch and using the trained encoder from 3D-OAE for fine-tuning. The results show that our proposed 3D-OAE can significantly accelerate convergence and improve accuracy. The training process of 3D-OAE is also more smooth and steady.
These results demonstrate that 3D-OAE can learn powerful representation from unlabelled data and transfer well to downstream tasks.


\begin{table}[tb]\small
\centering
\caption{\textbf{Ablation study on framework design.}}

\begin{tabular}{c|ccc|cc}
\toprule
Methods &Centralize &Loss & Occlusion &Linear Acc.  &Fine-t. Acc. \\ 
\midrule
Solution A & \XSolidBrush & CD & Rand & 20.8 & $-$ \\
Solution B & \Checkmark & EMD  & Rand & 88.4 & 92.4\\
Solution C & \Checkmark & CD  & Block & 91.5 & 92.7\\
Solution D & \Checkmark & CD & Rand & \textbf{92.3} & \textbf{93.4}\\
\bottomrule
\end{tabular}
\label{table:ablation1}
\end{table}

\subsection{Ablation study}
\label{section:sec4.7}
We analyze the effectiveness of each design in 3D-OAE. For convenience, we conduct all experiments on the ShapeNet dataset, and report the classification accuracy of both Linear SVM and supervised fine-tuning in ModelNet40. By default, all the experiment settings remains the same as Sec. \ref{section:sec4.1}, except for the analyzed part.

\subsubsection{Effect of each design in 3D-OAE}
To evaluate the effectiveness of each design in 3D-OAE, we make comparisons between four different experimental solutions shown in Table \ref{table:ablation1}. Solution A is trained without centralizing point patches to seed points. Solution B is trained using Earth Mover's Distance as the loss function. Solution C is trained with a block occlusion strategy. And Solution D is our default setting. It's clear that all the proposed designs in 3D-OAE can improve the performance of our method. And we find that using patch mix strategy \cite{yun2019cutmix,zhang2021pointcutmix} fails to enhance the representation learning ability of 3D-OAE.


\subsubsection{Occluding ratio}
Table \ref{table:ablation4} shows the numerical comparison of different occlusion ratios. with an occlusion ratio of 0, the auto-encoder fails to learn a powerful representation from the self-reconstruction task, which proves the effectiveness of our proposed occlusion strategy. Similar to MAE \cite{he2021masked}, we find that the occlusion ratio of 75$\%$ performs the best on both the linear accuracy and supervised fine-tuning accuracy. This is very different from BERT-style self-supervised learning works, where BRET masks only 15$\%$ of words and Point-BERT choose to occlude 25$\%$ to 45$\%$ of the point patches.
\begin{table}[t]\small
    \centering
    \setlength{\tabcolsep}{2mm}
    \caption{\textbf{Ablation study on occlusion ratios.} }
    \begin{tabular}{c|ccccc}
    \toprule
    Occlusion ratio  & 0 & 0.5 & 0.65 & \textbf{0.75} & 0.85 \\ 
    \midrule
    Linear Acc. & 59.2 & 90.9 & 91.1 & \textbf{92.3} & 90.7\\
    Fine-t. Acc. & 92.1 & 93.1 & 92.7 & \textbf{93.4} & 93.0\\
    \bottomrule
    \end{tabular}
    \label{table:ablation2}
\end{table}

\begin{table}[ht]\small
    \centering
    \setlength{\tabcolsep}{2mm}
    \caption{\textbf{Ablation study on group numbers.} }
    \begin{tabular}{c|cccc}
    \toprule
    Group  & 16 & 32 & \textbf{64} & 128 \\ 
    \midrule
    Linear Acc. & 88.6 & 91.1 & \textbf{92.3} & 91.4\\
    Fine-t. Acc. & 92.6 & 92.5 & \textbf{93.4} & 92.6\\
    \bottomrule
    \end{tabular}
    \label{table:ablation3}
\end{table}

\begin{table}[ht]\small
    \centering
    \setlength{\tabcolsep}{2mm}
    \caption{\textbf{Ablation study on the patch size.} }
    \begin{tabular}{c|cccc}
    \toprule
    Patch size  & 8 & 16 & \textbf{32} & 64 \\ 
    \midrule
    Linear Acc. & 75.6 & 90.4 & \textbf{92.3} & 91.5\\
    Fine-t. Acc. & 92.2 & 92.5 & \textbf{93.4} & 92.4\\
    \bottomrule
    \end{tabular}
    \label{table:ablation4}
\end{table}

\subsubsection{Grouping strategy}
Table \ref{table:ablation3} and \ref{table:ablation4} show the influence of the group numbers $G$ and the patch size $K$. It's clear that the chosen grouping strategy with a group number of 64 and a patch size of 32 achieves the best accuracy. For a fair comparison with other self-supervised methods, we adopt the same grouping strategy when reproducing them.

\section{Conclusion and Future Works}
In this paper, we present a novel point cloud self-supervised learning framework, named 3D Occlusion Auto-Encoder (3D-OAE). Our method shows powerful ability on transferring the learned representations to various downstream tasks, even in generative tasks and on real-world data. Specifically, we conduct comprehensive experiments on six different downstream tasks, and achieves SOTA performance in each task.
These results show that predicting complete shapes from highly occluded ones is an effective way of self-supervised representation learning for point clouds. Moreover, 3D-OAE also shows great efficiency since our encoder only operates on 25$\%$ of the input point cloud patches. 

Although our proposed 3D-OAE can learn powerful representations efficiently from the unlabelled point clouds, it still has some limitations which can be further improved in the future. For example, 3D-OAE separates the point cloud into patches and uses a self-reconstruction framework to learn patch-level representations. But the supervision is still constructed globally, which means that some local details cannot be fully considered. We consider introducing more local region constraints, such as conducting optimization losses between each local patch and its corresponding target.

\bibliographystyle{IEEEtran}
\bibliography{egbib}

\begin{thebibliography}{10}
\providecommand{\url}[1]{#1}
\csname url@samestyle\endcsname
\providecommand{\newblock}{\relax}
\providecommand{\bibinfo}[2]{#2}
\providecommand{\BIBentrySTDinterwordspacing}{\spaceskip=0pt\relax}
\providecommand{\BIBentryALTinterwordstretchfactor}{4}
\providecommand{\BIBentryALTinterwordspacing}{\spaceskip=\fontdimen2\font plus
\BIBentryALTinterwordstretchfactor\fontdimen3\font minus
  \fontdimen4\font\relax}
\providecommand{\BIBforeignlanguage}[2]{{%
\expandafter\ifx\csname l@#1\endcsname\relax
\typeout{** WARNING: IEEEtran.bst: No hyphenation pattern has been}%
\typeout{** loaded for the language `#1'. Using the pattern for}%
\typeout{** the default language instead.}%
\else
\language=\csname l@#1\endcsname
\fi
#2}}
\providecommand{\BIBdecl}{\relax}
\BIBdecl

\bibitem{cui2021deep}
Y.~Cui, R.~Chen, W.~Chu, L.~Chen, D.~Tian, Y.~Li, and D.~Cao, ``Deep learning
  for image and point cloud fusion in autonomous driving: A review,''
  \emph{IEEE Transactions on Intelligent Transportation Systems}, 2021.

\bibitem{alexiou2017towards}
E.~Alexiou, E.~Upenik, and T.~Ebrahimi, ``Towards subjective quality assessment
  of point cloud imaging in augmented reality,'' in \emph{2017 IEEE 19th
  International Workshop on Multimedia Signal Processing (MMSP)}.\hskip 1em
  plus 0.5em minus 0.4em\relax IEEE, 2017, pp. 1--6.

\bibitem{pomerleau2015review}
F.~Pomerleau, F.~Colas, and R.~Siegwart, ``A review of point cloud registration
  algorithms for mobile robotics,'' \emph{Foundations and Trends in Robotics},
  vol.~4, no.~1, pp. 1--104, 2015.

\bibitem{guo2021pct}
M.-H. Guo, J.-X. Cai, Z.-N. Liu, T.-J. Mu, R.~R. Martin, and S.-M. Hu, ``{PCT:
  Point cloud transformer},'' \emph{Computational Visual Media}, vol.~7, no.~2,
  pp. 187--199, 2021.

\bibitem{yu2021pointbert}
X.~Yu, L.~Tang, Y.~Rao, T.~Huang, J.~Zhou, and J.~Lu, ``Point-{BERT}:
  Pre-training {3D} point cloud transformers with masked point modeling,'' in
  \emph{Proceedings of the IEEE Conference on Computer Vision and Pattern
  Recognition (CVPR)}, 2022.

\bibitem{huang2021spatio}
S.~Huang, Y.~Xie, S.-C. Zhu, and Y.~Zhu, ``Spatio-temporal self-supervised
  representation learning for {3D} point clouds,'' in \emph{Proceedings of the
  IEEE/CVF International Conference on Computer Vision}, 2021, pp. 6535--6545.

\bibitem{he2021masked}
K.~He, X.~Chen, S.~Xie, Y.~Li, P.~Doll{\'a}r, and R.~Girshick, ``Masked
  autoencoders are scalable vision learners,'' in \emph{Proceedings of the
  IEEE/CVF Conference on Computer Vision and Pattern Recognition}, 2022, pp.
  16\,000--16\,009.

\bibitem{ma2021neural}
B.~Ma, Z.~Han, Y.-S. Liu, and M.~Zwicker, ``{Neural-Pull}: Learning signed
  distance functions from point clouds by learning to pull space onto
  surfaces,'' in \emph{Proceedings of the 38th International Conference on
  Machine Learning}, vol. 139, 2021.

\bibitem{chen2021unsupervised}
C.~Chen, Z.~Han, Y.-S. Liu, and M.~Zwicker, ``Unsupervised learning of fine
  structure generation for {3D} point clouds by 2d projections matching,'' in
  \emph{Proceedings of the IEEE/CVF International Conference on Computer
  Vision}, 2021, pp. 12\,466--12\,477.

\bibitem{li2022learning}
T.~Li, X.~Wen, Y.-S. Liu, H.~Su, and Z.~Han, ``Learning deep implicit functions
  for 3d shapes with dynamic code clouds,'' in \emph{Proceedings of the
  IEEE/CVF Conference on Computer Vision and Pattern Recognition}, 2022, pp.
  12\,840--12\,850.

\bibitem{3DAttriFlow}
X.~Wen, J.~Zhou, Y.-S. Liu, H.~Su, Z.~Dong, and Z.~Han, ``{3D} shape
  reconstruction from {2D} images with disentangled attribute flow,'' in
  \emph{Proceedings of the IEEE/CVF Conference on Computer Vision and Pattern
  Recognition (CVPR)}, 2022.

\bibitem{On-SurfacePriors}
B.~Ma, Y.-S. Liu, and Z.~Han, ``Reconstructing surfaces for sparse point clouds
  with on-surface priors,'' in \emph{Proceedings of the IEEE/CVF Conference on
  Computer Vision and Pattern Recognition}, 2022, pp. 6315--6325.

\bibitem{PredictableContextPrior}
B.~Ma, Y.-S. Liu, M.~Zwicker, and Z.~Han, ``Surface reconstruction from point
  clouds by learning predictive context priors,'' in \emph{Proceedings of the
  IEEE/CVF Conference on Computer Vision and Pattern Recognition}, 2022, pp.
  6326--6337.

\bibitem{spunet}
X.~Liu, X.~Liu, Y.-S. Liu, and Z.~Han, ``{SPU-Net}: Self-supervised point cloud
  upsampling by coarse-to-fine reconstruction with self-projection
  optimization,'' \emph{IEEE Transactions on Image Processing}, pp. 1--1, 2022.

\bibitem{pmp++}
X.~Wen, P.~Xiang, Z.~Han, Y.-P. Cao, P.~Wan, W.~Zheng, and Y.-S. Liu,
  ``{PMP-Net++}: Point cloud completion by transformer-enhanced multi-step
  point moving paths,'' \emph{IEEE Transactions on Pattern Analysis and Machine
  Intelligence}, pp. 1--1, 2022.

\bibitem{Zhou2022CAP-UDF}
J.~Zhou, B.~Ma, L.~Yu-Shen, F.~Yi, and H.~Zhizhong, ``Learning
  consistency-aware unsigned distance functions progressively from raw point
  clouds,'' in \emph{Advances in Neural Information Processing Systems
  (NeurIPS)}, 2022.

\bibitem{sheng2021deep}
X.~Sheng, L.~Li, D.~Liu, Z.~Xiong, Z.~Li, and F.~Wu, ``Deep-pcac: An end-to-end
  deep lossy compression framework for point cloud attributes,'' \emph{IEEE
  Transactions on Multimedia}, vol.~24, pp. 2617--2632, 2021.

\bibitem{akhtar2021video}
A.~Akhtar, W.~Gao, L.~Li, Z.~Li, W.~Jia, and S.~Liu, ``Video-based point cloud
  compression artifact removal,'' \emph{IEEE Transactions on Multimedia}, 2021.

\bibitem{qiu2021geometric}
S.~Qiu, S.~Anwar, and N.~Barnes, ``Geometric back-projection network for point
  cloud classification,'' \emph{IEEE Transactions on Multimedia}, vol.~24, pp.
  1943--1955, 2021.

\bibitem{qi2017pointnet}
C.~R. Qi, H.~Su, K.~Mo, and L.~J. Guibas, ``{PointNet: Deep learning on point
  sets for 3D classification and segmentation},'' in \emph{Proceedings of the
  IEEE/CVF Conference on Computer Vision and Pattern Recognition}, 2017, pp.
  652--660.

\bibitem{qi2017pointnet++}
C.~R. Qi, L.~Yi, H.~Su, and L.~J. Guibas, ``Point{Net}++: Deep hierarchical
  feature learning on point sets in a metric space,'' \emph{Advances in neural
  information processing systems}, vol.~30, 2017.

\bibitem{wu2019pointconv}
W.~Wu, Z.~Qi, and L.~Fuxin, ``{PointConv: Deep convolutional networks on 3D
  point clouds},'' in \emph{Proceedings of the IEEE/CVF Conference on Computer
  Vision and Pattern Recognition}, 2019, pp. 9621--9630.

\bibitem{yang2019modeling}
J.~Yang, Q.~Zhang, B.~Ni, L.~Li, J.~Liu, M.~Zhou, and Q.~Tian, ``Modeling point
  clouds with self-attention and gumbel subset sampling,'' in \emph{Proceedings
  of the IEEE/CVF Conference on Computer Vision and Pattern Recognition}, 2019,
  pp. 3323--3332.

\bibitem{hu2020randla}
Q.~Hu, B.~Yang, L.~Xie, S.~Rosa, Y.~Guo, Z.~Wang, N.~Trigoni, and A.~Markham,
  ``Randla-{N}et: Efficient semantic segmentation of large-scale point
  clouds,'' in \emph{Proceedings of the IEEE/CVF Conference on Computer Vision
  and Pattern Recognition}, 2020, pp. 11\,108--11\,117.

\bibitem{zhang2020pointhop}
M.~Zhang, H.~You, P.~Kadam, S.~Liu, and C.-C.~J. Kuo, ``Pointhop: An
  explainable machine learning method for point cloud classification,''
  \emph{IEEE Transactions on Multimedia}, vol.~22, no.~7, pp. 1744--1755, 2020.

\bibitem{wang2019dynamic}
Y.~Wang, Y.~Sun, Z.~Liu, S.~E. Sarma, M.~M. Bronstein, and J.~M. Solomon,
  ``Dynamic graph {CNN} for learning on point clouds,'' \emph{Acm Transactions
  On Graphics (ToG)}, vol.~38, no.~5, pp. 1--12, 2019.

\bibitem{gadelha2018multiresolution}
M.~Gadelha, R.~Wang, and S.~Maji, ``Multiresolution tree networks for {3D}
  point cloud processing,'' in \emph{Proceedings of the European Conference on
  Computer Vision}, 2018, pp. 103--118.

\bibitem{verma2018feastnet}
N.~Verma, E.~Boyer, and J.~Verbeek, ``{Feast{N}et: Feature-steered graph
  convolutions for 3D shape analysis},'' in \emph{Proceedings of the IEEE/CVF
  Conference on Computer Vision and Pattern Recognition}, 2018, pp. 2598--2606.

\bibitem{shen2018mining}
Y.~Shen, C.~Feng, Y.~Yang, and D.~Tian, ``Mining point cloud local structures
  by kernel correlation and graph pooling,'' in \emph{Proceedings of the
  IEEE/CVF Conference on Computer Vision and Pattern Recognition}, 2018, pp.
  4548--4557.

\bibitem{wang2019graph}
L.~Wang, Y.~Huang, Y.~Hou, S.~Zhang, and J.~Shan, ``Graph attention convolution
  for point cloud semantic segmentation,'' in \emph{Proceedings of the IEEE/CVF
  Conference on Computer Vision and Pattern Recognition}, 2019, pp.
  10\,296--10\,305.

\bibitem{chen2020hapgn}
C.~Chen, S.~Qian, Q.~Fang, and C.~Xu, ``Hapgn: Hierarchical attentive pooling
  graph network for point cloud segmentation,'' \emph{IEEE Transactions on
  Multimedia}, vol.~23, pp. 2335--2346, 2020.

\bibitem{hua2018pointwise}
B.-S. Hua, M.-K. Tran, and S.-K. Yeung, ``Pointwise convolutional neural
  networks,'' in \emph{Proceedings of the IEEE/CVF Conference on Computer
  Vision and Pattern Recognition}, 2018, pp. 984--993.

\bibitem{xu2018spidercnn}
Y.~Xu, T.~Fan, M.~Xu, L.~Zeng, and Y.~Qiao, ``Spider{CNN}: Deep learning on
  point sets with parameterized convolutional filters,'' in \emph{Proceedings
  of the European Conference on Computer Vision}, 2018, pp. 87--102.

\bibitem{zhang2019shellnet}
Z.~Zhang, B.-S. Hua, and S.-K. Yeung, ``Shell{N}et: Efficient point cloud
  convolutional neural networks using concentric shells statistics,'' in
  \emph{Proceedings of the IEEE/CVF International Conference on Computer
  Vision}, 2019, pp. 1607--1616.

\bibitem{su2018splatnet}
H.~Su, V.~Jampani, D.~Sun, S.~Maji, E.~Kalogerakis, M.-H. Yang, and J.~Kautz,
  ``{SplatNet}: Sparse lattice networks for point cloud processing,'' in
  \emph{Proceedings of the IEEE/CVF Conference on Computer Vision and Pattern
  Recognition}, 2018, pp. 2530--2539.

\bibitem{li2018pointcnn}
Y.~Li, R.~Bu, M.~Sun, W.~Wu, X.~Di, and B.~Chen, ``{PointCNN}: Convolution on
  x-transformed points,'' \emph{Advances in Neural Information Processing
  Systems}, vol.~31, 2018.

\bibitem{thomas2019kpconv}
H.~Thomas, C.~R. Qi, J.-E. Deschaud, B.~Marcotegui, F.~Goulette, and L.~J.
  Guibas, ``{KPConv}: Flexible and deformable convolution for point clouds,''
  in \emph{Proceedings of the IEEE/CVF International Conference on Computer
  Vision}, 2019, pp. 6411--6420.

\bibitem{liu2019point2sequence}
X.~Liu, Z.~Han, Y.-S. Liu, and M.~Zwicker, ``Point2{S}equence: Learning the
  shape representation of {3D} point clouds with an attention-based sequence to
  sequence network,'' in \emph{Proceedings of the AAAI Conference on Artificial
  Intelligence}, vol.~33, no.~01, 2019, pp. 8778--8785.

\bibitem{wen2020point2spatialcapsule}
X.~Wen, Z.~Han, X.~Liu, and Y.-S. Liu, ``Point2spatialcapsule: Aggregating
  features and spatial relationships of local regions on point clouds using
  spatial-aware capsules,'' \emph{IEEE Transactions on Image Processing},
  vol.~29, pp. 8855--8869, 2020.

\bibitem{liu2019relation}
Y.~Liu, B.~Fan, S.~Xiang, and C.~Pan, ``Relation-shape convolutional neural
  network for point cloud analysis,'' in \emph{Proceedings of the IEEE/CVF
  Conference on Computer Vision and Pattern Recognition}, 2019, pp. 8895--8904.

\bibitem{xu2020geometry}
M.~Xu, Z.~Zhou, and Y.~Qiao, ``Geometry sharing network for {3D} point cloud
  classification and segmentation,'' in \emph{Proceedings of the AAAI
  Conference on Artificial Intelligence}, vol.~34, no.~07, 2020, pp.
  12\,500--12\,507.

\bibitem{ma2021rethinking}
X.~Ma, C.~Qin, H.~You, H.~Ran, and Y.~Fu, ``Rethinking network design and local
  geometry in point cloud: A simple residual mlp framework,'' in
  \emph{International Conference on Learning Representations}, 2021.

\bibitem{vaswani2017attention}
A.~Vaswani, N.~Shazeer, N.~Parmar, J.~Uszkoreit, L.~Jones, A.~N. Gomez,
  {\L}.~Kaiser, and I.~Polosukhin, ``Attention is all you need,''
  \emph{Advances in Neural Information Processing Systems}, vol.~30, 2017.

\bibitem{devlin2018bert}
J.~Devlin, M.-W. Chang, K.~Lee, and K.~Toutanova, ``{BERT}: Pre-training of
  deep bidirectional transformers for language understanding,'' \emph{arXiv
  preprint arXiv:1810.04805}, 2018.

\bibitem{joshi2020spanbert}
M.~Joshi, D.~Chen, Y.~Liu, D.~S. Weld, L.~Zettlemoyer, and O.~Levy,
  ``Span{BERT}: Improving pre-training by representing and predicting spans,''
  \emph{Transactions of the Association for Computational Linguistics}, vol.~8,
  pp. 64--77, 2020.

\bibitem{carion2020end}
N.~Carion, F.~Massa, G.~Synnaeve, N.~Usunier, A.~Kirillov, and S.~Zagoruyko,
  ``End-to-end object detection with transformers,'' in \emph{Proceedings of
  the European Conference on Computer Vision}.\hskip 1em plus 0.5em minus
  0.4em\relax Springer, 2020, pp. 213--229.

\bibitem{liu2021swin}
Z.~Liu, Y.~Lin, Y.~Cao, H.~Hu, Y.~Wei, Z.~Zhang, S.~Lin, and B.~Guo, ``Swin
  {T}ransformer: Hierarchical vision transformer using shifted windows,'' in
  \emph{Proceedings of the IEEE/CVF International Conference on Computer
  Vision}, 2021, pp. 10\,012--10\,022.

\bibitem{dosovitskiy2020image}
A.~Dosovitskiy, L.~Beyer, A.~Kolesnikov, D.~Weissenborn, X.~Zhai,
  T.~Unterthiner, M.~Dehghani, M.~Minderer, G.~Heigold, S.~Gelly \emph{et~al.},
  ``An image is worth 16x16 words: Transformers for image recognition at
  scale,'' \emph{arXiv preprint arXiv:2010.11929}, 2020.

\bibitem{bao2021beit}
H.~Bao, L.~Dong, and F.~Wei, ``{BEiT}: Bert pre-training of image
  transformers,'' \emph{arXiv preprint arXiv:2106.08254}, 2021.

\bibitem{zhao2021point}
H.~Zhao, L.~Jiang, J.~Jia, P.~H. Torr, and V.~Koltun, ``Point transformer,'' in
  \emph{Proceedings of the IEEE/CVF International Conference on Computer
  Vision}, 2021, pp. 16\,259--16\,268.

\bibitem{yu2021pointr}
X.~Yu, Y.~Rao, Z.~Wang, Z.~Liu, J.~Lu, and J.~Zhou, ``{PoinTr: Diverse point
  cloud completion with geometry-aware transformers},'' in \emph{Proceedings of
  the IEEE/CVF International Conference on Computer Vision}, 2021, pp.
  12\,498--12\,507.

\bibitem{han2022dual}
X.-F. Han, Y.-F. Jin, H.-X. Cheng, and G.-Q. Xiao, ``Dual transformer for point
  cloud analysis,'' \emph{IEEE Transactions on Multimedia}, 2022.

\bibitem{yang2018foldingnet}
Y.~Yang, C.~Feng, Y.~Shen, and D.~Tian, ``Folding{Net}: Point cloud
  auto-encoder via deep grid deformation,'' in \emph{Proceedings of the
  IEEE/CVF Conference on Computer Vision and Pattern Recognition}, 2018, pp.
  206--215.

\bibitem{xie2020pointcontrast}
S.~Xie, J.~Gu, D.~Guo, C.~R. Qi, L.~Guibas, and O.~Litany, ``{PointContrast:
  Unsupervised pre-training for 3D point cloud understanding},'' in
  \emph{Proceedings of the European Conference on Computer Vision}.\hskip 1em
  plus 0.5em minus 0.4em\relax Springer, 2020, pp. 574--591.

\bibitem{wang2021unsupervised}
H.~Wang, Q.~Liu, X.~Yue, J.~Lasenby, and M.~J. Kusner, ``Unsupervised point
  cloud pre-training via occlusion completion,'' in \emph{Proceedings of the
  IEEE/CVF International Conference on Computer Vision}, 2021, pp. 9782--9792.

\bibitem{sun2021point}
C.~Sun, Z.~Zheng, X.~Wang, M.~Xu, and Y.~Yang, ``Point cloud pre-training by
  mixing and disentangling,'' \emph{arXiv preprint arXiv:2109.00452}, 2021.

\bibitem{sauder2019self}
J.~Sauder and B.~Sievers, ``Self-supervised deep learning on point clouds by
  reconstructing space,'' \emph{Advances in Neural Information Processing
  Systems}, vol.~32, 2019.

\bibitem{sanghi2020info3d}
A.~Sanghi, ``{Info3D: Representation learning on 3D objects using mutual
  information maximization and contrastive learning},'' in \emph{Proceedings of
  the European Conference on Computer Vision}.\hskip 1em plus 0.5em minus
  0.4em\relax Springer, 2020, pp. 626--642.

\bibitem{rao2020global}
Y.~Rao, J.~Lu, and J.~Zhou, ``Global-local bidirectional reasoning for
  unsupervised representation learning of {3D} point clouds,'' in
  \emph{Proceedings of the IEEE/CVF Conference on Computer Vision and Pattern
  Recognition}, 2020, pp. 5376--5385.

\bibitem{li2018so}
J.~Li, B.~M. Chen, and G.~H. Lee, ``{SO-Net}: Self-organizing network for point
  cloud analysis,'' in \emph{Proceedings of the IEEE/CVF Conference on Computer
  Vision and Pattern Recognition}, 2018, pp. 9397--9406.

\bibitem{eckart2021self}
B.~Eckart, W.~Yuan, C.~Liu, and J.~Kautz, ``Self-supervised learning on {3D}
  point clouds by learning discrete generative models,'' in \emph{Proceedings
  of the IEEE/CVF Conference on Computer Vision and Pattern Recognition}, 2021,
  pp. 8248--8257.

\bibitem{zhang2021self}
Z.~Zhang, R.~Girdhar, A.~Joulin, and I.~Misra, ``Self-supervised pretraining of
  {3D} features on any point-cloud,'' in \emph{Proceedings of the IEEE/CVF
  International Conference on Computer Vision}, 2021, pp. 10\,252--10\,263.

\bibitem{afham2022crosspoint}
M.~Afham, I.~Dissanayake, D.~Dissanayake, A.~Dharmasiri, K.~Thilakarathna, and
  R.~Rodrigo, ``Cross{P}oint: Self-supervised cross-modal contrastive learning
  for {3D} point cloud understanding,'' in \emph{IEEE/CVF International
  Conference on Computer Vision and Pattern Recognition}, June 2022.

\bibitem{han2019multi}
Z.~Han, X.~Wang, Y.-S. Liu, and M.~Zwicker, ``{Multi-Angle Point Cloud-VAE:
  Unsupervised feature learning for 3D point clouds from multiple angles by
  joint self-reconstruction and half-to-half prediction},'' in
  \emph{Proceedings of the IEEE/CVF International Conference on Computer
  Vision}.\hskip 1em plus 0.5em minus 0.4em\relax IEEE, 2019, pp.
  10\,441--10\,450.

\bibitem{liu2019l2g}
X.~Liu, Z.~Han, X.~Wen, Y.-S. Liu, and M.~Zwicker, ``{L2G} auto-encoder:
  Understanding point clouds by local-to-global reconstruction with
  hierarchical self-attention,'' in \emph{Proceedings of the 27th ACM
  International Conference on Multimedia}, 2019, pp. 989--997.

\bibitem{yan2022implicit}
S.~Yan, Z.~Yang, H.~Li, L.~Guan, H.~Kang, G.~Hua, and Q.~Huang, ``Implicit
  autoencoder for point cloud self-supervised representation learning,''
  \emph{arXiv preprint arXiv:2201.00785}, 2022.

\bibitem{chang2015shapenet}
A.~X. Chang, T.~Funkhouser, L.~Guibas, P.~Hanrahan, Q.~Huang, Z.~Li,
  S.~Savarese, M.~Savva, S.~Song, H.~Su \emph{et~al.}, ``{ShapeNet: An
  information-rich 3D model repository},'' \emph{arXiv preprint
  arXiv:1512.03012}, 2015.

\bibitem{loshchilov2017decoupled}
I.~Loshchilov and F.~Hutter, ``Decoupled weight decay regularization,''
  \emph{arXiv preprint arXiv:1711.05101}, 2017.

\bibitem{zhao20193d}
Y.~Zhao, T.~Birdal, H.~Deng, and F.~Tombari, ``{3D} point capsule networks,''
  in \emph{Proceedings of the IEEE/CVF Conference on Computer Vision and
  Pattern Recognition}, 2019, pp. 1009--1018.

\bibitem{wu20153d}
Z.~Wu, S.~Song, A.~Khosla, F.~Yu, L.~Zhang, X.~Tang, and J.~Xiao, ``{3D
  ShapeNets: A deep representation for volumetric shapes},'' in
  \emph{Proceedings of the IEEE/CVF Conference on Computer Vision and Pattern
  Recognition}, 2015, pp. 1912--1920.

\bibitem{wu2016learning}
J.~Wu, C.~Zhang, T.~Xue, B.~Freeman, and J.~Tenenbaum, ``Learning a
  probabilistic latent space of object shapes via {3D} generative-adversarial
  modeling,'' \emph{Advances in Neural Information Processing Systems},
  vol.~29, 2016.

\bibitem{han2019view}
Z.~Han, M.~Shang, Y.-S. Liu, and M.~Zwicker, ``{View inter-prediction GAN:
  Unsupervised representation learning for 3D shapes by learning global shape
  memories to support local view predictions},'' in \emph{Proceedings of the
  AAAI Conference on Artificial Intelligence}, vol.~33, no.~01, 2019, pp.
  8376--8384.

\bibitem{valsesia2020learning}
D.~Valsesia, G.~Fracastoro, and E.~Magli, ``Learning localized representations
  of point clouds with graph-convolutional generative adversarial networks,''
  \emph{IEEE Transactions on Multimedia}, vol.~23, pp. 402--414, 2020.

\bibitem{poursaeed2020self}
O.~Poursaeed, T.~Jiang, H.~Qiao, N.~Xu, and V.~G. Kim, ``Self-supervised
  learning of point clouds via orientation estimation,'' in \emph{2020
  International Conference on 3D Vision (3DV)}.\hskip 1em plus 0.5em minus
  0.4em\relax IEEE, 2020, pp. 1018--1028.

\bibitem{van2008visualizing}
L.~Van~der Maaten and G.~Hinton, ``Visualizing data using {t-SNE}.''
  \emph{Journal of machine learning research}, vol.~9, no.~11, 2008.

\bibitem{sharma2020self}
C.~Sharma and M.~Kaul, ``Self-supervised few-shot learning on point clouds,''
  \emph{Advances in Neural Information Processing Systems}, vol.~33, pp.
  7212--7221, 2020.

\bibitem{yi2016scalable}
L.~Yi, V.~G. Kim, D.~Ceylan, I.-C. Shen, M.~Yan, H.~Su, C.~Lu, Q.~Huang,
  A.~Sheffer, and L.~Guibas, ``A scalable active framework for region
  annotation in {3D} shape collections,'' \emph{ACM Transactions on Graphics
  (ToG)}, vol.~35, no.~6, pp. 1--12, 2016.

\bibitem{yuan2018pcn}
W.~Yuan, T.~Khot, D.~Held, C.~Mertz, and M.~Hebert, ``{PCN}: Point completion
  network,'' in \emph{2018 International Conference on 3D Vision (3DV)}.\hskip
  1em plus 0.5em minus 0.4em\relax IEEE, 2018, pp. 728--737.

\bibitem{tchapmi2019topnet}
L.~P. Tchapmi, V.~Kosaraju, H.~Rezatofighi, I.~Reid, and S.~Savarese,
  ``{TopNet}: {S}tructural point cloud decoder,'' in \emph{Proceedings of the
  IEEE/CVF Conference on Computer Vision and Pattern Recognition}, 2019, pp.
  383--392.

\bibitem{groueix2018papier}
T.~Groueix, M.~Fisher, V.~G. Kim, B.~C. Russell, and M.~Aubry, ``A
  papier-m{\^a}ch{\'e} approach to learning 3{D} surface generation,'' in
  \emph{Proceedings of the IEEE/CVF Conference on Computer Vision and Pattern
  Recognition}, 2018, pp. 216--224.

\bibitem{xie2020grnet}
H.~Xie, H.~Yao, S.~Zhou, J.~Mao, S.~Zhang, and W.~Sun, ``{GRNet}: Gridding
  residual network for dense point cloud completion,'' in \emph{Proceedings of
  the European Conference on Computer Vision}.\hskip 1em plus 0.5em minus
  0.4em\relax Springer, 2020, pp. 365--381.

\bibitem{wang2020cascaded}
X.~Wang, M.~H. Ang~Jr, and G.~H. Lee, ``Cascaded refinement network for point
  cloud completion,'' in \emph{Proceedings of the IEEE/CVF Conference on
  Computer Vision and Pattern Recognition}, 2020, pp. 790--799.

\bibitem{wen2021pmp}
X.~Wen, P.~Xiang, Z.~Han, Y.-P. Cao, P.~Wan, W.~Zheng, and Y.-S. Liu,
  ``{PMP-Net: Point cloud completion by learning multi-step point moving
  paths},'' in \emph{Proceedings of the IEEE/CVF Conference on Computer Vision
  and Pattern Recognition}, 2021, pp. 7443--7452.

\bibitem{zhang2020detail}
W.~Zhang, Q.~Yan, and C.~Xiao, ``Detail preserved point cloud completion via
  separated feature aggregation,'' in \emph{Proceedings of the European
  Conference on Computer Vision}.\hskip 1em plus 0.5em minus 0.4em\relax
  Springer, 2020, pp. 512--528.

\bibitem{xiang2021snowflakenet}
P.~Xiang, X.~Wen, Y.-S. Liu, Y.-P. Cao, P.~Wan, W.~Zheng, and Z.~Han,
  ``{SnowflakeNet: Point cloud completion by snowflake point deconvolution with
  skip-transformer},'' in \emph{Proceedings of the IEEE/CVF International
  Conference on Computer Vision}, 2021, pp. 5499--5509.

\bibitem{uy2019revisiting}
M.~A. Uy, Q.-H. Pham, B.-S. Hua, T.~Nguyen, and S.-K. Yeung, ``Revisiting point
  cloud classification: A new benchmark dataset and classification model on
  real-world data,'' in \emph{Proceedings of the IEEE/CVF International
  Conference on Computer Vision}, 2019, pp. 1588--1597.

\bibitem{yun2019cutmix}
S.~Yun, D.~Han, S.~J. Oh, S.~Chun, J.~Choe, and Y.~Yoo, ``Cutmix:
  Regularization strategy to train strong classifiers with localizable
  features,'' in \emph{Proceedings of the IEEE/CVF International Conference on
  Computer Vision}, 2019, pp. 6023--6032.

\bibitem{zhang2021pointcutmix}
J.~Zhang, L.~Chen, B.~Ouyang, B.~Liu, J.~Zhu, Y.~Chen, Y.~Meng, and D.~Wu,
  ``{PointCutmix}: Regularization strategy for point cloud classification,''
  \emph{arXiv preprint arXiv:2101.01461}, 2021.

\end{thebibliography}

\vskip 0pt plus -1fil
\begin{IEEEbiography}[{\includegraphics[width=1in,height=1.25in,clip,keepaspectratio]{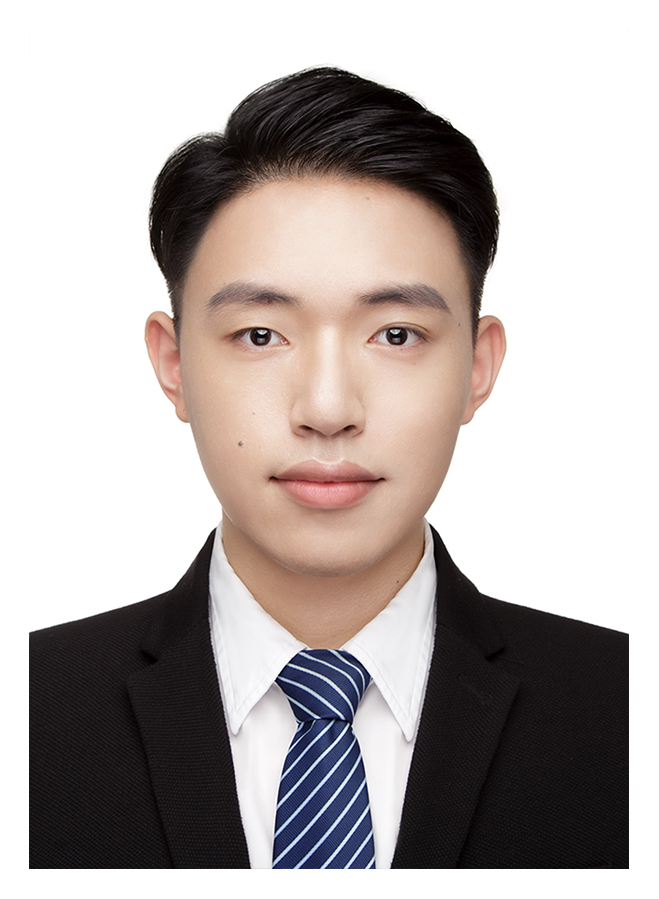}}]{Junsheng Zhou}
    received the B.S. degree in software engineering from Xiamen University, China, in 2021. He is currently the graduate student with the School of Software, Tsinghua University. His research interests include deep learning, 3D shape analysis and 3D reconstruction.
\end{IEEEbiography}

\begin{IEEEbiography}[{\includegraphics[width=1in,height=1.25in,clip,keepaspectratio]{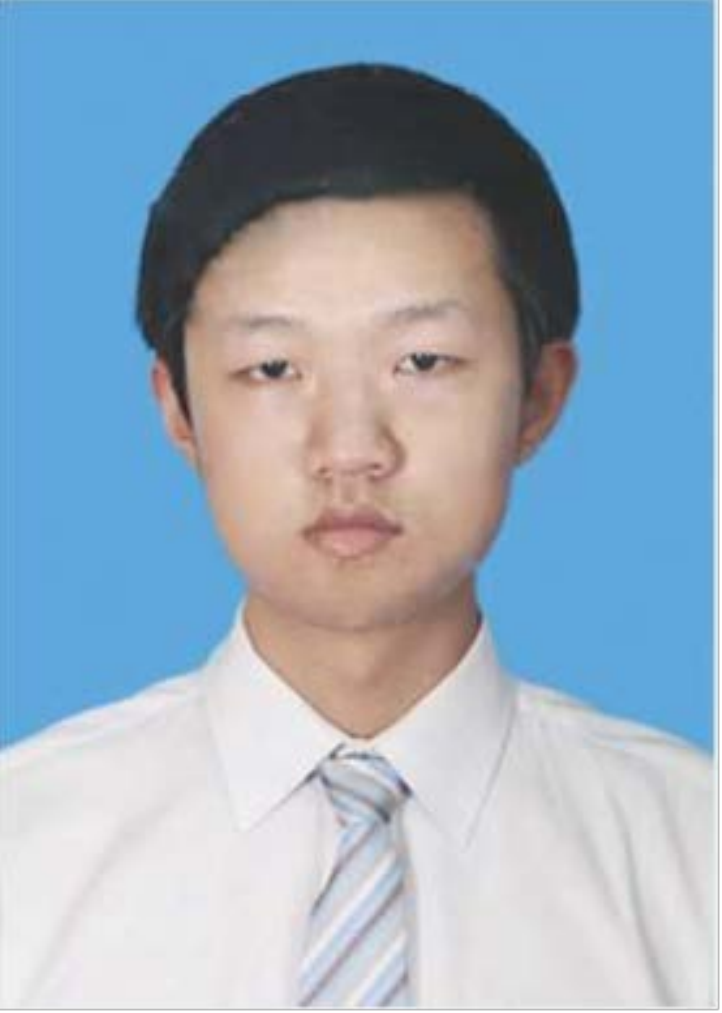}}]{Xin Wen}
    received the Ph.D. degree from the School of software Tsinghua University, China, in 2021. He is currently a research engineer at JD Logistics.  His research interests include deep learning, shape analysis and pattern recognition, and NLP.
\end{IEEEbiography}

\begin{IEEEbiography}[{\includegraphics[width=1in,height=1.25in,clip,keepaspectratio]{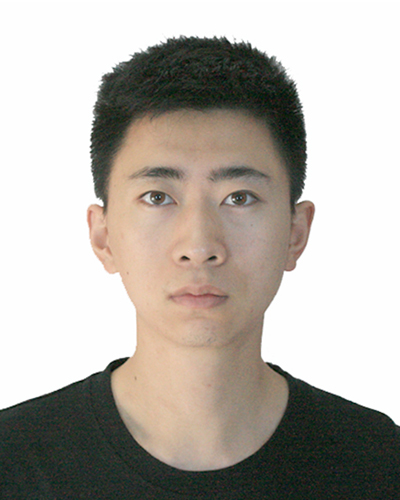}}]{Baorui Ma}
    received the B.S. degree in computer science and technology from Jilin University, China, in 2018. He is currently pursuing the Ph.D. degree with the School of Software, Tsinghua University. His research interests include deep learning and 3D reconstruction.
\end{IEEEbiography}

\begin{IEEEbiography}[{\includegraphics[width=1in,height=1.25in,clip,keepaspectratio]{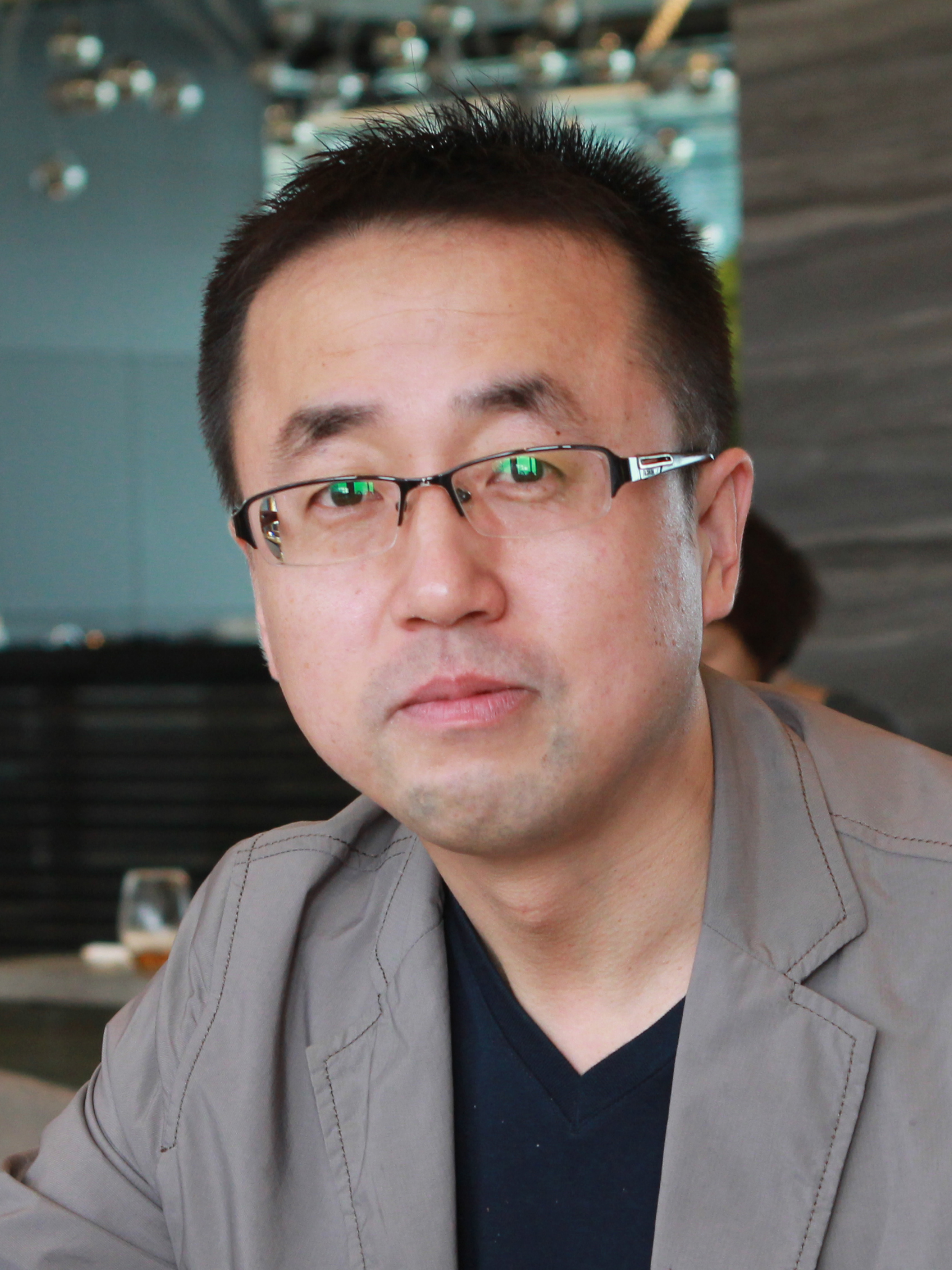}}]{Yu-Shen Liu}
    (M'18) received the B.S. degree in mathematics from Jilin University, China, in 2000, and the Ph.D. degree from the Department of Computer Science and Technology, Tsinghua University, Beijing, China, in 2006. From 2006 to 2009, he was a Post-Doctoral Researcher with Purdue University. He is currently an Associate Professor with the School of Software, Tsinghua University. His research interests include shape analysis, pattern recognition, machine learning, and semantic search.
\end{IEEEbiography}

\begin{IEEEbiography}[{\includegraphics[width=1in,height=1.25in,clip,keepaspectratio]{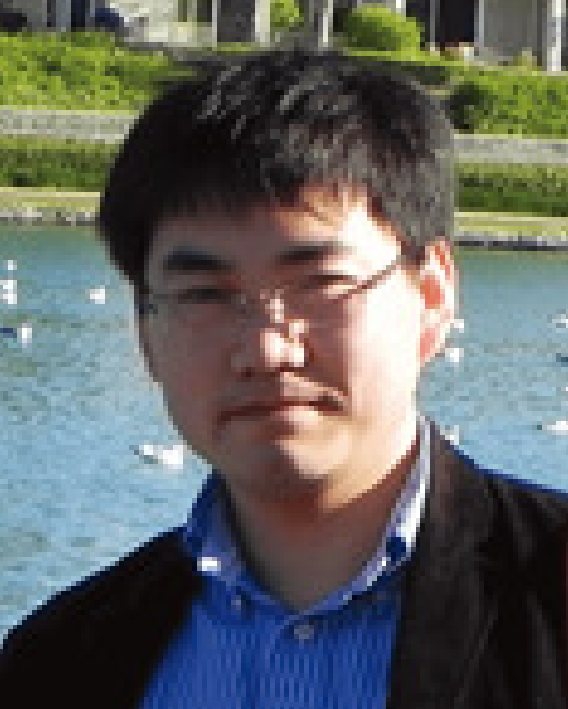}}]{Yue Gao}
   (SM'14)  is an associate professor with the School of Software, Tsinghua University. He received the B.S. degree from the Harbin Institute of Technology, Harbin, China, and the M.E. and Ph.D. degrees from Tsinghua University, Beijing, China. From 2012 to 2014, he was a research fellow in National University of Singapore. From 2014 to 2016, he was a postdoctral research associate in University of North Carolina, Chapel Hill. His research interests include on artificial intelligence and computer vision.
\end{IEEEbiography}


\begin{IEEEbiography}[{\includegraphics[width=1in,height=1.25in,clip]{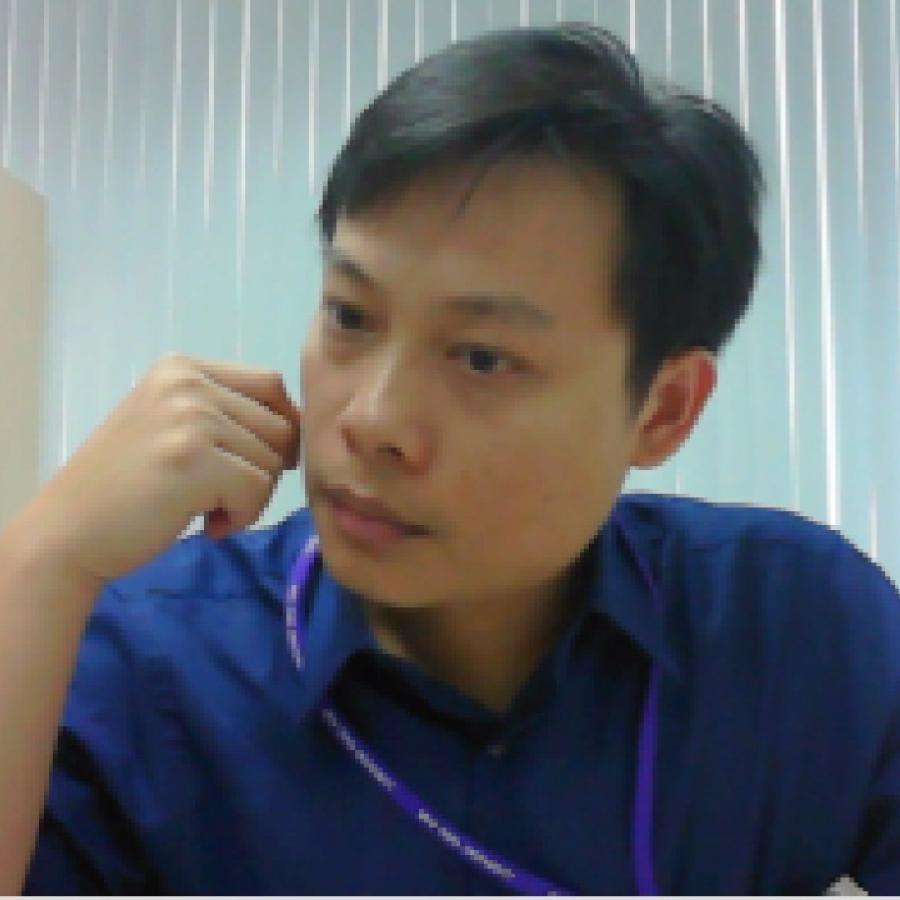}}]{Yi Fang}
     received B.S. and M.S. degrees in biomedical engineering from Xian Jiaotong University, Xian, China, in 2003 and 2006, respectively, and a Ph.D. in mechanical engineering from Purdue University, West Lafayette, IN, USA, in 2011. He is currently an Associate Professor with the Department of Electrical and Computer Engineering, New York University Abu Dhabi, Abu Dhabi, United Arab Emirates. His research interests include 3D computer vision and pattern recognition, large-scale visual computing, deep visual computing, deep cross-domain and crossmodality multimedia analysis, and computational structural biology.
\end{IEEEbiography}

\begin{IEEEbiography}[{\includegraphics[width=1in,height=1.25in,clip,keepaspectratio]{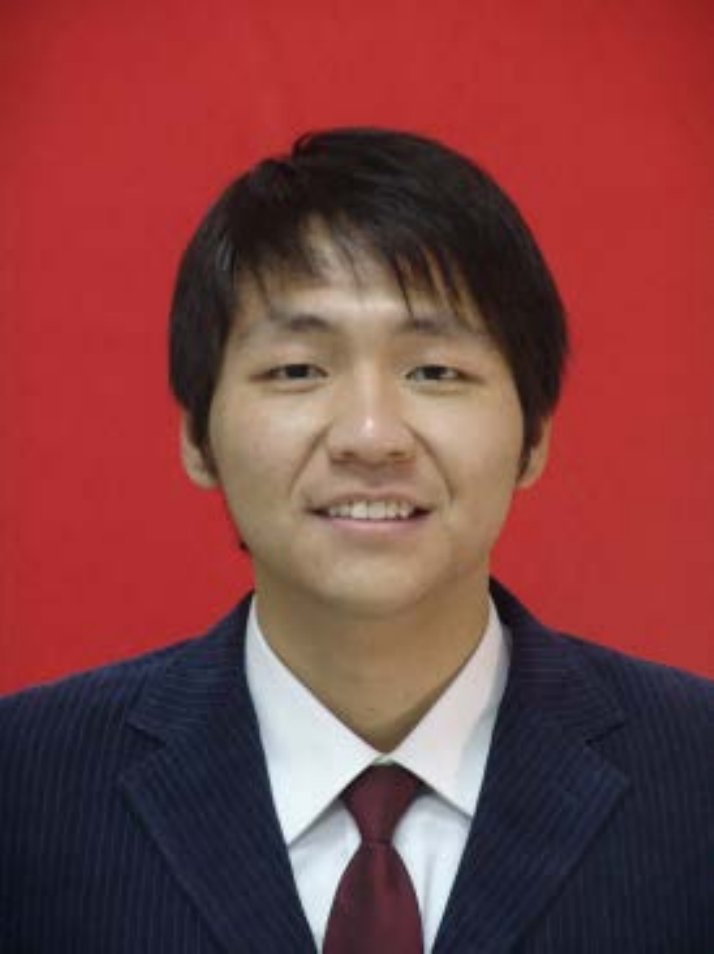}}]{Zhizhong Han}
    received the Ph.D. degree from Northwestern Polytechnical University, China, 2017. He was a Post-Doctoral Researcher with the Department of Computer Science, at the University of Maryland, College Park, USA. Currently, he is an Assistant Professor of Computer Science at Wayne State University, USA. His research interests include 3D computer vision, digital geometry processing and artificial intelligence.
\end{IEEEbiography}

\vfill

\end{document}